\documentclass[11pt]{article}

\usepackage[final]{acl}

\usepackage{times}
\usepackage{latexsym}

\usepackage[T1]{fontenc}

\usepackage[utf8]{inputenc}

\usepackage{microtype}

\usepackage{inconsolata}

\usepackage{graphicx}

\usepackage{wrapfig}

\usepackage{amsmath}
\usepackage{amssymb} 
\usepackage{cleveref}

\usepackage{graphicx}
\usepackage{caption}
\usepackage{subcaption}
\usepackage{algorithm}
\usepackage{algpseudocode} %
\usepackage{booktabs}
\usepackage{tabularx}
\usepackage{dblfloatfix}
\usepackage{array}
\usepackage{float}
\usepackage{enumitem}

\usepackage{placeins} 
\usepackage[russian,english]{babel}       
\usepackage{tabularx,array,ragged2e}
\newcolumntype{Y}{>{\RaggedRight\arraybackslash}X} 

\title{Small Data, Big Noise: Adversarial Training for Robust Parameter-Efficient Fine-Tuning}

  \author{
  Eitan Cohen \\
  Bar-Ilan University \\
  Ramat Gan, Israel \\
  \texttt{ceitan001@gmail.com}
  \And
  Idan Simai \\
  Bar-Ilan University \\
  Ramat Gan, Israel \\
  \texttt{idansi98@gmail.com}
  \And
  Uri Shaham \\
  Bar-Ilan University \\
  Ramat Gan, Israel \\
  \texttt{uri.shaham@biu.ac.il}
}

\begin{document}
\maketitle

\begin{abstract}
Parameter-Efficient Fine-Tuning (PEFT) has become essential for adapting foundation models to downstream NLP tasks. However, current PEFT methods often struggle with robustness to noise and performance degradation on limited training data. We propose \textbf{SDBN} (Small Data Big Noise), a unified framework that brings adversarial training to PEFT - a combination that remains less studied in the PEFT setting despite its complementary strengths - to enhance model robustness and generalization, outperforming alternative approaches. We also introduce two variants of the method that use discrete uncertainty sets: \textbf{SDBN-h}, which enumerates character-level edits and selects worst-case variants using gradients, and \textbf{SDBN-p}, which uses LLM-generated variants for robust optimization in generative tasks. Experiments across multiple benchmarks reveal substantial improvements, particularly in low-resource settings and under both word-level and character-level corruptions. This framework addresses the less explored intersection of adversarial training and parameter-efficient adaptation, without introducing additional parameters or only modest computational overhead, making PEFT deployments more reliable in real-world scenarios where data scarcity and linguistic variability often coexist.
\end{abstract}

\section{Introduction}
\label{intro}

Parameter-Efficient Fine-Tuning (PEFT) has recently emerged as a promising strategy for adapting Large Language Models (LLMs) to downstream tasks, while substantially reducing both computational and storage requirements. Notable PEFT approaches include Adapter \cite{adapter}, BitFit \cite{bitfit}, and Low-Rank Adaptation (LoRA) \cite{lora}. In particular, LoRA has garnered significant attention, spurring the development of several variants \cite{melora}, \cite{adalora},\cite{qlora} which further extend its applicability.
\begin{figure}[ht]
      \includegraphics[width=\linewidth]{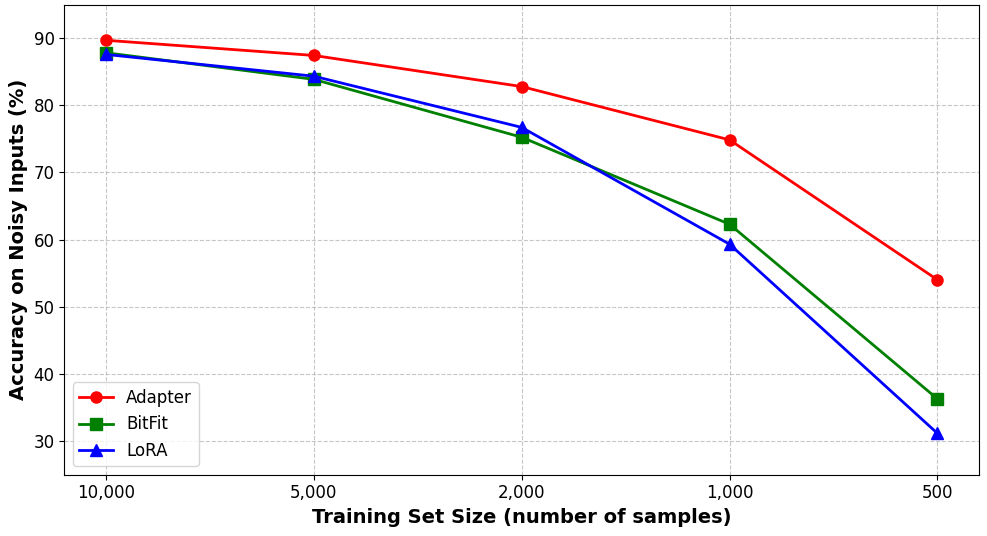}
      \caption{\textbf{Performance degradation of PEFT methods on limited training data.} Accuracy of three common PEFT methods: Adapter, BitFit, and LoRA--on a test set with \emph{word‑swapping} noise as the size of the \textit{clean} Banking77 training data corpus is reduced (x‑axis, number of examples). All methods exhibit a marked drop in noisy‑test accuracy as the available training data \textit{dwindles}, in some cases losing more than 50{\%} below 1,000 training samples. The trend underscores the current vulnerability of PEFT models to realistic textual perturbations in low‑resource settings.}
      \label{fig:noise_scaling}
\end{figure}

While these techniques substantially reduce the computational overhead required for fine-tuning foundation models, they frequently underperform when confronted with domain shifts and noisy data (see \cref{fig:noise_scaling})  - conditions that commonly appear in practical NLP deployments. Real-world text often contains imperfections such as typos, inconsistent formatting, and dialectal variations that can significantly degrade model performance. Additionally, while PEFT approaches have demonstrated impressive results on large-scale training corpora, their effectiveness may deteriorate significantly when training samples are limited - a common scenario in domains such as medical data, aerospace, extinct languages, and similar fields. Notably, \cite{bitfit} shows that PEFT methods can be more efficient than full fine-tuning on small datasets, motivating a focused investigation of PEFT approaches tailored specifically for settings with limited data. These challenges--robustness to noise and domain shifts, as well as worse performance in low-resource settings--remain largely underexplored in current PEFT methodologies.

Motivated by these challenges, we propose SDBN (Small Data Big Noise)\footnote{\textbf{Code:} \url{https://github.com/shaham-lab/SDBN}}, a unified framework that integrates adversarial training principles into PEFT to enhance robustness and generalization capabilities, including robustness to tokenization-breaking \emph{character-level} corruptions in addition to word-level noise. While adversarial training has been widely adopted in various contexts \cite{freelb,smart,advlora}--primarily for defense against attacks--its application to improving PEFT methods in low-resource, noisy settings remains unexplored. The proposed SDBN framework improves model performance under both noisy conditions (word- and character-level) and limited training data scenarios. Beyond continuous embedding-space perturbations, we introduce two complementary strategies for constructing discrete uncertainty sets: SDBN-h, which employs gradient-guided character-level edits to address tokenization-breaking noise, and SDBN-p, which leverages LLM-generated adversarial variants for richer semantic perturbations particularly suited to generative tasks. These discrete uncertainty-set instantiations replace norm ball uncertainty sets in regimes it cannot cover - tokenization-breaking character edits (SDBN-h) and semantic, generation-oriented variants (SDBN-p).\ Unlike most PEFT approaches - which do not explicitly target scenarios with restricted data or address domain shifts - SDBN achieves substantial improvements across various benchmark datasets. Notably, this framework provides robustness not only to known perturbations but also to unanticipated domain shifts that may emerge during deployment, without requiring domain-specific adaptation data. Importantly, SDBN maintains the parameter efficiency of existing PEFT methods without adding trainable parameters or extra GPU memory overhead, offering a practical solution for deploying robust language models in resource-constrained and noisy real-world environments.

Our Contributions are as follows: First, We propose \textbf{SDBN}, a framework that brings adversarial training to PEFT - a combination on \textbf{less studied in the PEFT setting} - to improve robustness in low-resource settings. Second, we introduce two variants of the method utilizing discrete uncertainty sets: \textbf{SDBN-h} based on character-level perturbation, and \textbf{SDBN-p} based on LLM-generated perturbations.
Third, we demonstrate \textbf{substantial robustness gains} across multiple benchmarks under word-level and character-level corruptions, without additional parameters.

\section{Related work}

\paragraph{PEFT methods.} Recent advances in PEFT methods-such as LoRA, Adapter, BitFit, Prompt \cite{prompt} and Prefix \cite{prefix}--have facilitated the efficient adaptation of pre-trained models by fine-tuning only a limited set of parameters. In particular, LoRA employs low-rank updates, Adapter incorporates trainable bottleneck modules within each layer, and BitFit restricts modifications to bias parameters. Notably, recent evaluations demonstrate that PEFT methods like LoRA actually exhibit superior robustness to textual noise and adversarial perturbations compared to full fine-tuning \cite{rusert2025investigating}. However, even with this inherent advantage, these models remain vulnerable to realistic corruptions in data-scarce regimes, providing a strong foundation for the targeted robust-optimization mechanisms introduced in SDBN. While these approaches reduce parameter counts, most standard PEFT implementations were not originally designed with robustness to noisy inputs \cite{TowardsRobustAndGeneralizedPEFT} or data-limited scenarios as primary optimization targets. Our work explores how integrating adversarial training techniques with existing PEFT methods can address these limitations, particularly in low-resource settings. 

\paragraph{Robust optimization.}
Robust optimization is widely recognized as an effective strategy for enhancing model stability in the presence of noisy data and domain shifts. \cite{SHAHAM2018195}, \cite{madry2019deep} demonstrated that robust optimization can significantly improve a model's generalization capabilities on noisy data and downstream tasks, using neural networks; however, the robust optimization approaches discussed in their works, which directly inject noise into raw input data, require specific adaptations for the NLP domain and were not evaluated within PEFT frameworks. \cite{TowardsRobustAndGeneralizedPEFT} applied robust optimization to address noisy labels but did not consider noisy input data. Moreover, neither approach focused on optimization under conditions of limited data. In contrast, the integration of adversarial training with PEFT frameworks explored in this work addresses both noisy inputs and small datasets simultaneously.

\paragraph{Adversarial Training.} Using adversarial training is one way to achieve robust optimization. FreeLB \cite{freelb}, SMART \cite{smart} and VAT \cite{miyato2016adversarial} applied adversarial training to LLMs, primarily focusing on improving generalization capabilities; however, those do not incorporate PEFT methodologies, which can render full fine-tuning impractical in certain scenarios. Recent methods such as LoFT \cite{loft} and AdvLoRA \cite{advlora} apply adversarial training with LoRA on vision tasks (in the case of AdvLoRA, mainly on the visual component of vision-language tasks), leaving their applicability in NLP domains unexplored. Notably, the effectiveness of adversarial training with PEFT methods in scenarios with small datasets and noisy textual inputs remains largely uninvestigated. This work examines how adversarial training can be applied to PEFT paradigms in NLP and demonstrates enhanced performance over conventional PEFT baselines, particularly in low-resource settings.

\paragraph{PEFT in NLP under low resources.} PEFT is now the de‑facto strategy for adapting large language models to downstream tasks, yet its behaviour under noisy or limited data remains understudied. NEFtune
\cite{neftune} shows that injecting \emph{uniform random} noise into token embeddings during instruction tuning (NEFTune) markedly improves \emph{average‑case} performance, focusing primarily on enhancing model generalization, but they do not analyse worst‑case perturbations or data‑scarce regimes.
In contrast, this work applies \emph{adversarial} training to PEFT with a different objective: adding adversarial perturbations in the embedding space and optimising adapters to minimise the resulting worst‑case loss.
This targeted approach yields robustness to input noise and domain shift specifically in scenarios with small, noisy datasets--precisely the conditions where PEFT methods are most practically attractive but often struggle without additional robustness mechanisms.

\paragraph{Robustness through data noising.} Several works have shown that even small character edits (e.g., missing letters) can cause degradation in NLP models, since such changes often break tokenization and push inputs far in embedding space. EDA~\cite{eda} proposed simple data augmentation editing actions, showing improvements on small datasets; however, these perturbations are applied randomly rather than guided by gradients. HotFlip~\cite{hotflip} introduced gradient-guided character-level edits, but as an attack tool. WildNLP~\cite{wildnlp} and \citet{dialectnoise} explored noise-augmented training to improve robustness to character corruptions, but without leveraging adversarial signals to identify worst-case perturbations. In contrast, our work adapts character-level perturbations within an adversarial training framework, using gradient guidance to maximize their effectiveness for improving robustness in low-resource PEFT settings.

\paragraph{Discrete uncertainty sets.}
Several prior works use discrete perturbation sets in NLP, but in settings that differ substantially from ours.
Some focus on certified robustness via verification-style pipelines such as randomized masking or interval bound propagation, e.g., \citep{zeng-etal-2023-certified}, \citep{huang-etal-2019-achieving}, and \citep{jia-etal-2019-certified}.
These methods target formal guarantees against bounded substitution attacks rather than training-time robust optimization for PEFT, and are not naturally scalable to large generative models or practical PEFT training loops.
Other work, such as \citep{zhou-etal-2021-defense}, proposes an inference-time defense against synonym substitution attacks rather than adversarial training.
Closest in spirit is \citep{ivgi-berant-2021-achieving}, which studies discrete adversarial training for classification; however, it assumes full fine-tuning, focuses on attack-style substitutions rather than practical small-data noise, and does not consider PEFT or generative tasks.
In contrast, our work studies PEFT as the adaptation regime, uses discrete uncertainty sets within a unified training-time robust optimization framework, and evaluates them across multiple PEFT methods in low-resource noisy settings, including generative tasks.

\section{Background on robust optimization}
\label{adv_train_pre}
Robust optimization \citep{ben2009robust} is a field in optimization theory that aims to improve model stability under input uncertainty. Consider a model with parameters \(\theta\) and a dataset \(\mathcal{D}\) of input-label pairs \((x,y)\), where \(x \in \mathbb{R}^d\) and \(y \in \{1, \ldots, K\}\). For each input \(x\) (in our context- as a sentence), we define an \emph{uncertainty set}  \(\mathcal{S}_x \subseteq \mathbb{R}^d\): which reflects the level of uncertainty in the input- the set of sentences that arise from small, semantics‑preserving edits to \(x\) (e.g., token deletions, swaps, or character flips); these variants capture exactly the inputs on which the model’s classification is uncertain. Given a loss function \(\mathcal{L}\), the robust optimization objective is:

\begin{equation*}\label{eq:robust_opt}
\min_{\theta}\, \mathbb{E}_{(x,y) \sim \mathcal{D}} \!\Bigl[\max_{x+\delta \in \mathcal{S}_x} \mathcal{L}(x+\delta;\theta,y)\Bigr].
\end{equation*}

When holding \(\theta\) and $y$ fixed and viewing \(\mathcal{L}(x;\theta,y)\) as a function of $x$ we occasionally write \(\mathcal{L}_{\theta,y}(x)\).
The inner maximization problem aims to find a worst-case example of a given $x$ in the uncertainty set that achieves the highest loss. Using a first-order Taylor approximation around the input \(x\), we can express the loss on the perturbed example as:
\begin{equation}
\label{eq:taylor}
\mathcal{L}_{\theta,y}(x+\delta) \approx \mathcal{L}_{\theta,y}(x) + \langle \nabla \mathcal{L}_{\theta,y}(x), \delta \rangle.
\end{equation}
The optimal perturbation \(\delta^*\) for each training example is then:
\begin{equation}
\label{eq:optimal_delta}
\delta^* = \arg\max_{\delta:x+\delta \in S_x}  \langle \nabla \mathcal{L}_{\theta,y}(x), \delta \rangle.
\end{equation}
One way to define \(\mathcal{S}_x\) is by a norm ball centered at \(x\) with a small radius \(\epsilon\):
\begin{equation}
\label{eq:Sx}
\mathcal{S}_x = \{x+\delta \in \mathbb{R}^d : \|\delta\|_p \le \epsilon\}.
\end{equation}
\Cref{eq:taylor} explains why the perturbation \(\tilde{\delta}\)\footnote{Details on how the choice of \(p\) shapes \(\delta^*\) (e.g., \(\ell_\infty\), \(\ell_2\), \(\ell_1\)) and its connections to standard procedures like FGSM are deferred to Appendix~\ref{app:norms}.} increases the loss compared to the original input. For small perturbation (with small $\epsilon$) the perturbed $x$ is positively correlated with the direction of the gradient $g = \nabla \mathcal{L}_{\theta,y}(x)$ (i.e., their angle is less than \(90^{\circ}\)), making the inner product \(\langle g, \tilde{\delta}\rangle\) positive and thus increasing the loss. This strategy yields more challenging training examples and can improve robustness to domain shifts and noisy inputs.

\textbf{Adversarial Training} \cite{goodfellow-etal-2015-iclr} stands as a prominent defense strategy for enhancing model robustness against attacks. It uses adversarial examples, which are perturbed versions of the original points. In the context of robust optimization, this approach leverages approximated worst-case examples which define an uncertainty set $S_x$ around each input $x$, as detailed in Section~\ref{adv_train_pre}.

\section{Methodology}
In this section, we describe how the proposed SDBN framework integrates adversarial training techniques with PEFT methods to address the challenges of noise robustness and domain shifts under limited data resources. While these techniques have been previously explored for full model fine-tuning, our focus is on demonstrating their particular value when applied to PEFT methods in low-resource, noisy settings.
Conceptually, SDBN is a single robust-optimization framework instantiated with three uncertainty sets:
(i) standard $\ell_\infty$ ball,
(ii) discrete tokenization-breaking character edits (SDBN-h),
and (iii) discrete LLM-generated adversarial variants (SDBN-p).
We detail each variant in the following subsections. 
\begin{figure*}[t]
  \centering
  \setlength{\tabcolsep}{2pt}
  \begin{tabular}{cc}
    \begin{subfigure}[t]{0.48\linewidth}
      \centering
      \includegraphics[width=\linewidth]{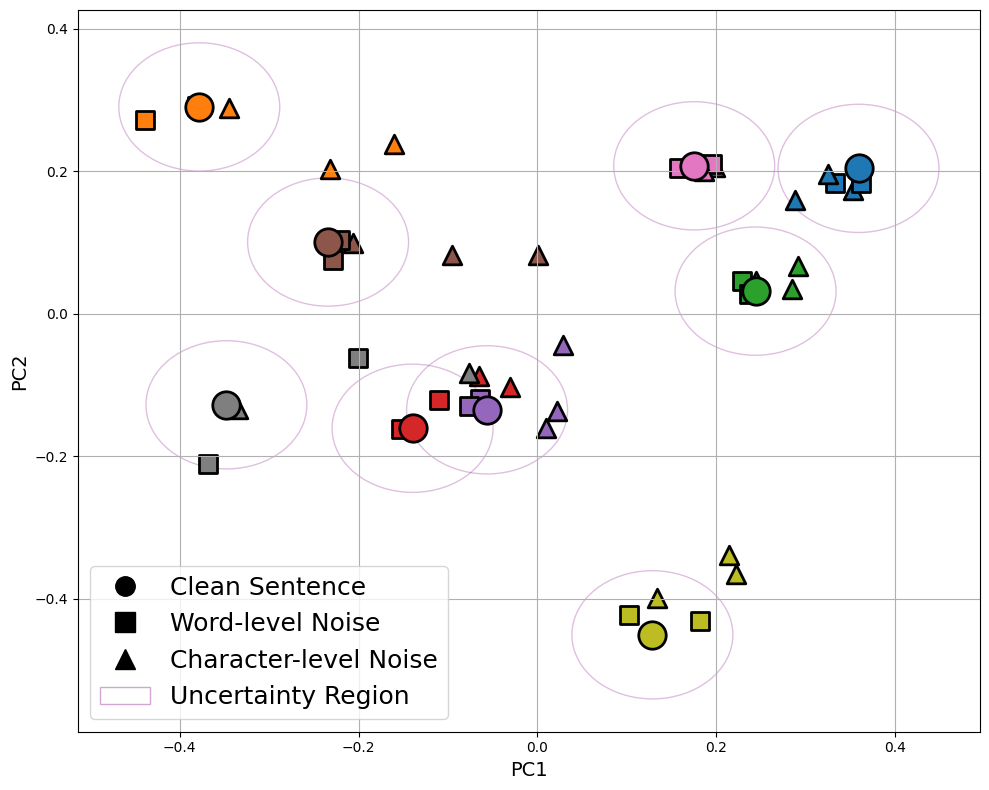}
      \caption{Character/lexical edits (\textit{TREC})}
      \label{fig:trec_noise}
    \end{subfigure}
    &
    \begin{subfigure}[t]{0.48\linewidth}
      \centering
      \includegraphics[width=\linewidth]{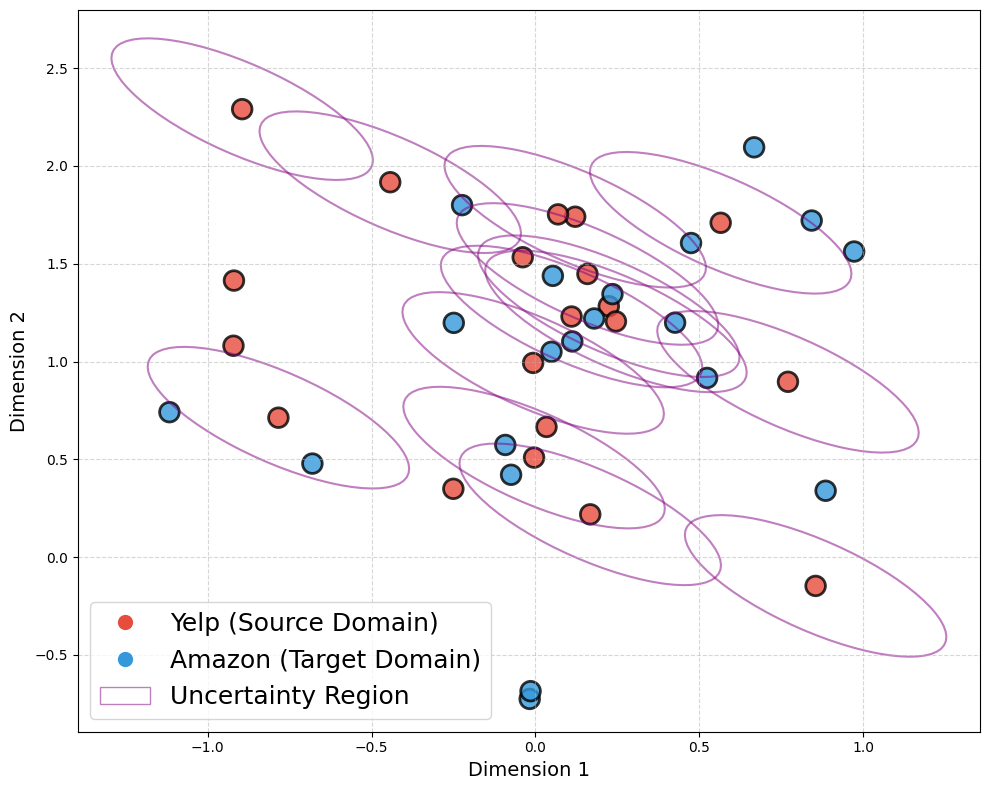}
      \caption{Domain Shift (\textit{$Yelp\!\rightarrow\!$Amazon})}
      \label{fig:tsne_domain}
    \end{subfigure}
  \end{tabular}
  \caption{\textbf{Perturbation regimes addressed by Adversarial Training.}
    (a) PCA visualization of embedding space for TREC sentences, showing how character/word  perturbations (triangles/squares) - fall within uncertainty regions (ellipses) around clean sentences (circles). This illustrates how our adversarial training approach effectively can improve the model's performance on diverse linguistic variations that occur within these uncertainty regions. Importantly, even when specific noisy examples are not encountered during training, the model develops robustness to these perturbations because they reside in the adversarially expanded uncertainty regions.
    (b) t-SNE plot of embeddings from Amazon and Yelp sentences, demonstrating a domain shift scenario where the uncertainty regions (ellipses) created around samples from the source domain (Yelp) encompass examples from the unseen target domain (Amazon). This visualization supports adversarial training key finding: by optimizing for worst-case examples within the uncertainty set during training, adversarial training generalizes effectively to novel domains without explicitly training on them.}
  \label{fig:noise_and_shift}
\end{figure*}

\subsection{Motivation}\label{sec}
We address two practical robustness challenges encountered in real-world NLP deployments, particularly in the prevalent low-resource scenarios where limited training data makes these issues even more severe. The first concerns fine-grained textual perturbations that occur naturally in user-generated content but are often absent from clean training data. These include various character and word-level modifications that preserve semantic meaning while changing the visible text structure. The second involves domain shifts that occur when deployment environments differ subtly from training conditions in style or topic distribution. Unlike traditional domain adaptation scenarios where target domain data is available, we aim to build resilience against unanticipated shifts without prior exposure to the target domain. The data scarcity compounds these challenges, as models trained on small datasets tend to overfit and lack the broad exposure needed to generalize well to variations. Our objective is to enhance PEFT methods to withstand these everyday linguistic variations even when trained on limited data, maintaining high performance on clean inputs while creating more reliable models for practical applications.

\subsection{Rationale}
\label{sec::rationale}
Motivated by \cite{SHAHAM2018195} and \cite{madry2019deep}, who demonstrate that adversarial training functions as a robust optimization procedure, the integration of PEFT with robust optimization through adversarial training addresses challenges specific to low-resource settings. As illustrated in Figure \ref{fig:noise_and_shift}, we can conceptualize both character/lexical variations and domain shifts as points within uncertainty regions surrounding clean examples. The PCA visualization in Figure \ref{fig:noise_and_shift}(a) demonstrates how character and word-level perturbations fall within uncertainty regions surrounding their clean sentence counterparts. Similarly, Figure \ref{fig:noise_and_shift}(b) shows how uncertainty regions created around samples from the source domain (Yelp) extend to cover examples from unseen target domain (Amazon).

Adversarial training optimizes robustness by training on \emph{worst-case} perturbations within the uncertainty region. This forces stability and yields generalization to both noisy inputs and unseen domains without explicit exposure, allowing the model to handle linguistic variations beyond the training set.

\textbf{Limited data scenarios.}
PEFT can outperform full fine-tuning on small datasets \cite{bitfit,Pu2023Empirical}, but limited corpora lack diversity in corruptions, dialects, and domain-specific terminology, making models \emph{sensitive} to noise and domain shift.
Adversarial training mitigates this by generating gradient-aligned \emph{worst-case} neighbors around each example and optimizing performance \emph{within the existing uncertainty set}.
Training on these hard variants builds robustness to unseen linguistic variations and domain differences—without additional labeled data. This intuition is also consistent with our empirical findings in results~\ref{peft_vs_full_ft}, where adversarial training yields substantially larger gains for PEFT methods than for full fine-tuning in the same low-resource setting under noisy conditions.

\textbf{Theoretical advantages of gradient-based perturbations over random noise.} The effectiveness of adversarial training compared to random noise injection can be formally understood through the lens of high-dimensional geometry and its effect on the optimization landscape. While both approaches introduce perturbations during training, their impact on model robustness differs fundamentally.

As established in \cref{eq:taylor}, the change in loss when applying a perturbation $\delta$ to the input is approximated by $\langle \nabla \mathcal{L}_{\theta,y}(x), \delta \rangle$. In high-dimensional embedding spaces, random noise vectors (as used in methods like \textsc{NEFTune} and DAE \cite{DAE}) exhibit a critical limitation: they are approximately orthogonal to any fixed direction, including the gradient. This is a well-established property in high-dimensional spaces as explained by \cite{miyato2016adversarial}, where random vectors become nearly perpendicular with high probability as dimensionality increases. Consequently, for random perturbations $\delta_{\text{random}}$ with constraint $\|\delta_{\text{random}}\|_p  \leq \varepsilon$, the expected inner product $\mathbb{E}[\langle \nabla \mathcal{L}_{\theta,y}(x), \delta_{\text{random}}\rangle] \approx 0$. This results in minimal consistent effect on the loss, creating only weak, undirected regularization that fails to target the model's specific vulnerabilities.

In contrast, adversarial perturbations as defined in \cref{eq:Sx} explicitly maximize the inner product with the gradient. The optimal perturbation $\delta^{\star}$ from \cref{eq:taylor} ensures the largest possible first-order increase in loss within the constraint. This targeted approach creates worst-case examples that probe precisely the directions where the model is most sensitive, compelling optimization to flatten the loss landscape exactly where it is steepest. While random noise merely induces general smoothing across all directions, adversarial training systematically expands decision margins in the regions that most require robustness, producing models that generalize effectively to both natural perturbations and domain shifts as visualized in Figure \ref{fig:noise_and_shift}.

\subsection{SDBN: PEFT with Adversarial Training}
In attempting to achieve robustness to noise using adversarial training, it is impossible to create adversarial examples by adding numeric perturbation to text symbols as in vision tasks (examining \cref{eq:Sx}, an incompatibility exists in the expression \(x+\delta\), where $x$ represents a sequence of discrete symbols while $\delta$ denotes a continuous numeric perturbation). Following established approaches in NLP adversarial training \cite{freelb},\cite{miyato2016adversarial},\cite{smart}, noise injection at the embedding layer is utilized rather than perturbing the raw input data.
Let \(E(\cdot)\) be the embedding extractor and \(f(\cdot;\theta)\) be the subsequent layers of the integrated model with any PEFT method (e.g., LoRA) and \(\theta\) denote the model's parameters. For an input batch \((X, Y)\), we compute the embeddings \(\mathbf{e} = E(X)\) and the clean loss:
\(
\mathcal{L}_{\text{clean}} = \mathcal{L}\bigl(f_\theta(\mathbf{e}), Y\bigr).
\)
We then compute its gradient \(\mathbf{g} = \nabla_{\mathbf{e}}\mathcal{L}_{\text{clean}}\) and form a perturbation:
\(
\delta = \epsilon \cdot \operatorname{sign}(\mathbf{g}),
\)
yielding adversarial embeddings \(\mathbf{e}_{\text{adv}} = \mathbf{e} + \delta\) and corresponding loss \(\mathcal{L}_{\text{adv}} = \mathcal{L}\bigl(f_{\theta}(\mathbf{e}_{\text{adv}}), Y\bigr)\). Finally, we update the trainable parameters \(\theta\) according to the specific PEFT method we use by \(\mathcal{L}_{\text{adv}}\) (for implementation details, see \cref{setup_details}).

Within the SDBN framework, this perturbation can be selected from any norm ball. Empirically, choosing from the \(\ell_{\infty}\) norm ball yields the best performance and is used for the experiments in \cref{sec:results}. For a detailed comparison of perturbations drawn from \(\ell_1\), \(\ell_2\), and \(\ell_\infty\) norm balls, see \cref{norms_method_appendix}. For further empirical motivation and visual intuition showing why $\ell_{\infty}$ with $\epsilon=10^{-4}$ best captures realistic noise patterns in embedding space, see the detailed analysis in \cref{intution}. For the description of epsilon selection and the pseudo-code, see \cref{adv_training_l_inf,epsilon_slection}.

\subsection{Character–level noise as a challenge}
As \cref{fig:trec_noise} indicates, most \emph{word}-level edits (squares) stay within the uncertainty region around clean sentences (circles), whereas many \emph{character}-level edits (triangles) fall outside it. 
Edits that \emph{break a token} - e.g., deleting a letter -alter tokenization (splits/\textsc{unk}) and push embeddings far from regions seen in training. By contrast, case changes typically preserve both tokenization (the word remains a single token) and semantics, so the resulting embedding stays close to the original. 
Appendix~\ref{break_token} provides concrete examples of this distance gap. 
Hence, robustness to character-level noise is more challenging and may require complementary tools alongside embedding-space adversarial training.

\textbf{Hybrid Strategy: SDBN-h.}
To address character-level perturbations that break tokenization and produce embeddings far outside the $\ell_p$-ball used in SDBN, we define a discrete uncertainty set $\mathcal{S}_x$ as all single-character variants of a sentence $x$. Since $\mathcal{S}_x$ is finite, projected gradient descent cannot be applied directly. Instead, given the gradient $g = \nabla_e \mathcal{L}(f_\theta(e), y)$ from the clean embedding $e=E(x)$, we select $z^*$ to be $\arg\max_{z \in \mathcal{S}_x} \langle g, E(z) - E(x) \rangle$ and use it as the adversarial example. Meaning, we use the gradient for selecting the perturbation but not for creating the perturbation. Each mini-batch is split: one subset is perturbed via standard $\ell_\infty$ FGSM in embedding space, and the other via $z^\star$, reusing the same gradient. This yields robustness to both continuous embedding perturbations and discrete character distortions with negligible overhead. For more details and full pseudo-code, see \cref{sdbn-h_appendix}. SDBN-h extends embedding-space adversarial training to tokenization-discrete character corruptions by performing a discrete worst-case selection using the same gradient signal.

\subsection{Prompt-Based Uncertainty Sets: SDBN-p} 
We have empirically found that the continuous embedding-space uncertainty set in SDBN is less effective for generative tasks. To tackle this, we construct an alternative discrete uncertainty set by leveraging an LLM to generate semantic-preserving adversarial variants. Given a training example $x$, we prompt an LLM to generate $k$ semantically-equivalent variants that include realistic perturbations such as paraphrases, typos, and style variations. Formally, we define $S_x^{\text{prompt}} = \{z_1, \ldots, z_k\}$, where each $z_i$ is generated by prompting an LLM with $x$ and instructions to produce adversarial variants. 

Unlike SDBN and SDBN-h, where perturbations are small enough to be guided by clean-input gradients via Taylor approximation, the variants in $S_x^{\text{prompt}}$ may involve significant structural changes that fall outside the local linear region. Consequently, for SDBN-p, we do not use the gradient-guided selection rule. Instead, we explicitly compute the loss for each of the $k$ pre-computed variants and select the one that maximizes the training objective: 
\begin{equation}
z^{\star} = \arg\max_{z \in S_x^{\text{prompt}}} \mathcal{L}(f(E(z); \theta), y)
\end{equation}
While this requires $k$ forward passes, it ensures we identify the true worst-case semantic variant for robust optimization in generative tasks. This approach naturally captures linguistic variations that may be difficult to enumerate algorithmically, such as paraphrases and contextual rewrites, making it particularly suitable for generative tasks where output diversity is important. SDBN-p uses an LLM-generated discrete uncertainty set instead of a norm-ball uncertainty set. For more details, examples, and pseudo-code, see \cref{sdbn-p_appendix}.

\section{Results}
\label{sec:results}
\textbf{Experimental Setup.}\label{sec:setup} We evaluate the proposed \textbf{SDBN} framework using two pre-trained models: BERT-base \cite{Devlin2019BERT} and DeBERTa-v3 \cite{He2023DeBERTaV3} across multiple classification benchmarks: sentences datasets including \textsc{20Newsgroups}~\citep{news20}, \textsc{Banking77}~\citep{banking77}, \textsc{TREC}~\citep{trec}, and \textsc{IMDB}, as well as the word-pair semantic relation classification dataset \textsc{BLESS}~\citep{bless}. We additionally evaluate on generative tasks across \textsc{SQuAD}~\citep{squad} and \textsc{TweetQA}~\citep{xiong-etal-2019-tweetqa} datasets, using LLaMA-3.2-1B~\citep{llama3herd}, LLaMA-2-7B~\citep{touvron2023llama2} ,and Qwen-2.5-7B~\citep{qwen2025qwen25} to assess robustness in generation-oriented tasks beyond classification. For SDBN-p, we use GPT-5.2~\citep{openai_gpt52_systemcard} to generate adversarial variants. For PEFT methods, we use: \textbf{Adapter}, \textbf{BitFit}, \textbf{LoRA}, and \textbf{QLoRA}.
We use SDBN ($\ell_\infty$ uncertainty set) as the default, SDBN-h for tokenization-breaking character noise,
and SDBN-p for generative tasks via precomputed LLM variant sets.

Our training protocol consists of 3 warm-up epochs with standard training followed by 10--20 epochs of either SDBN or baseline training without perturbations. We also compare to NEFTune (which was originally presented with QLoRA) and EDA (see implementation details in~\ref{app::eda})methods integrated with these PEFT approaches.

To simulate real-world scenarios, we evaluate under challenging conditions:
\textbf{Data scarcity:} using 5\% to 100\% of the original training data.
\textbf{Input noise:} applying word and char-level noise (for noise types details see \cref{noise_types}) at test time.

The remainder of our experimental evaluation follows a systematic progression: We assess performance on clean test data across varying training set sizes to establish baseline effectiveness and robustness under word/character corruptions to demonstrate the advantages of the SDBN framework in noisy environments. We then examine cross-domain generalization using the ArSarcasm-v2 \cite{abufarha-etal-2021-arsarcasm-v2} dataset and an NLI setting (cross-genre) to test performance under domain shifts. Due to space limits, all domain-shift results are reported in the Appendix (see ~\ref{app:domain-shift}). Across these domain-shift evaluations, our adversarially trained PEFT approach (SDBN) maintains consistent gains over baselines. Finally, we conduct targeted analyses of the SDBN's components, comparing different perturbation strategies and examining the impact of perturbation location within the model architecture, and reporting resource costs such as runtime and memory footprint (see ~\cref{sec:method_analysis}).

\begin{figure}[!tbp] 
  \centering
  \captionsetup{width=\columnwidth} 
  \setlength{\textfloatsep}{8pt plus 2pt minus 2pt} 
  \includegraphics[width=\columnwidth]{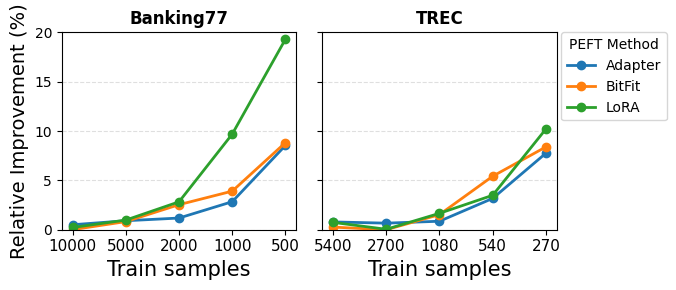}
  \caption{\textbf{Adversarial Training Effect on Limited Data Scenarios.}
  Relative performance improvement of adversarial training over baseline PEFT methods
  (e.g., 10\% means SDBN-LoRA/LoRA = 1.10) across training sizes on \textsc{Banking77}
  and \textsc{TREC}. Adversarial training not only maintains but often improves clean-data
  performance, with gains increasing as the training set shrinks.}
  \label{fig:clean_scales}
\end{figure}

\textbf{Results on clean data.}
\label{sec:clean_comparison} We evaluate models trained with standard PEFT methods alongside their SDBN variants on clean test sets. Our experiments demonstrate that SDBN consistently improves accuracy over vanilla PEFT methods across all datasets with limited training data. This confirms our theoretical analysis from Section~\ref{sec::rationale} that SDBN enhances generalization on clean data, not just under noisy conditions. Figure~\ref{fig:clean_scales} illustrates the relative improvements on \textsc{Banking77} and \textsc{TREC}, revealing a critical insight: the benefits of SDBN become increasingly significant as training data size decreases. This validates our hypothesis that robust optimization techniques are particularly valuable in low-resource scenarios, where models typically struggle with generalization. Full results across all data scales and datasets are in~\ref{results_clean}.

\begin{figure*}[!t] 
    
  \centering
  \captionsetup{width=\textwidth}
  \includegraphics[width=\textwidth]{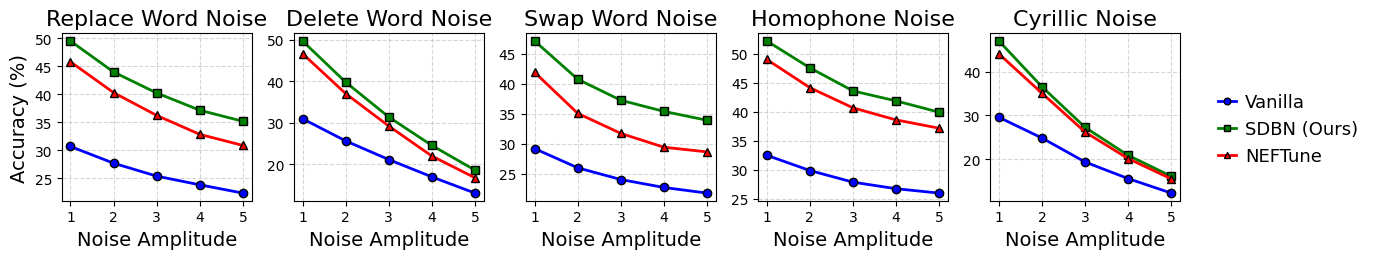}
  \caption{\textbf{Adversarial training for variable-intensity noise.}
  Performance comparison of LoRA PEFT implementations under variable-intensity noise with DeBERTa-v3 on 1{,}000 \textsc{Banking77} samples.
 SDBN shows superior robustness in low-resource settings.}
  \label{fig:noise_amp}
\end{figure*}

\newcolumntype{Y}{>{\raggedright\arraybackslash}p{0.55\textwidth}}
\newcolumntype{Y}{>{\raggedright\arraybackslash}X}
\newcolumntype{C}{>{\centering\arraybackslash}p{0.09\textwidth}} 
\textbf{Results on Noisy Data.}\label{tab:results} We evaluate model robustness to various linguistic perturbations using a 1,000-sample subset of \textsc{Banking77} with DeBERTa-v3, focusing on low-resource scenarios. As illustrated in Figure~\ref{fig:noise_amp} for variable-intensity noise, where we vary the perturbation amplitude from 1-5 operations per sentence (e.g., deleting one word vs. five words), SDBN maintains consistent advantage over both vanilla PEFT and NEFTune. The performance gap remains stable as corruption severity increases, showing SDBN's superior ability to handle progressively more distorted inputs. Results for constant-intensity noise, where each sentence is corrupted by exactly one operation, show similar improvements and are provided in Appendix~\ref{app::noise_const}.

These results validate our theoretical framework from \cref{sec::rationale}, where we represented linguistic variations as points within uncertainty regions surrounding clean examples. By optimizing for worst-case examples within these regions through gradient-based perturbations, SDBN effectively improves the model's robustness to real-world text variations, as visualized in \cref{fig:trec_noise}. In contrast, NEFTune's random noise approach, while helpful, lacks the targeted nature of adversarial training. SDBN's systematic exploration of uncertainty regions enables more effective generalization to diverse perturbations without explicit exposure during training, making it particularly valuable for low-resource scenarios where the limited training data lacks natural linguistic diversity.
\begin{table}[!t]
  \centering
  \captionsetup{width=\columnwidth}
  \caption{\textbf{BLESS: accuracy (\%) under character-level noise.}
DeBERTa-v3 + LoRA trained on 1{,}000 clean samples.
SDBN-h yields the best robustness on tokenization-breaking types.}
  \label{tab:sdbn_h_results}
  \setlength{\tabcolsep}{6pt}
  \renewcommand{\arraystretch}{1.15}
  \subcaptionbox{Clean, Delete-char, Swap-char\label{tab:sdbn_h_a}}{%
    \footnotesize
    \begin{tabular}{lccc}
      \toprule
      \textbf{Method} & \textbf{Clean} & \textbf{Delete-char} & \textbf{Swap-char} \\
      \midrule
      Vanilla    & 89.81 $\pm$ 0.29 & 60.67 $\pm$ 2.19 & 56.82 $\pm$ 2.38 \\
      SDBN    & \textbf{89.83 $\pm$ 0.24} & 60.84 $\pm$ 1.85 & 57.22 $\pm$ 2.84 \\
      NEFTune & 89.08 $\pm$ 0.54 & 61.19 $\pm$ 0.89 & 57.22 $\pm$ 0.82 \\
      SDBN-h  & 89.61 $\pm$ 0.30 & \textbf{65.14 $\pm$ 1.19} & \textbf{62.80 $\pm$ 1.47} \\
      \bottomrule
    \end{tabular}
  }
  \subcaptionbox{Double-char\label{tab:sdbn_h_b}}{%
    \footnotesize
    \begin{tabular}{lc}
      \toprule
      \textbf{Method} & \textbf{Double-char} \\
      \midrule
      Vanilla    & 68.51 $\pm$ 2.75 \\
      SDBN    & 68.66 $\pm$ 2.27 \\
      NEFTune & 69.42 $\pm$ 1.29 \\
      SDBN-h  & \textbf{72.54 $\pm$ 1.05} \\
      \bottomrule
    \end{tabular}
  }
\end{table}

\textbf{Character Noise (SDBN-h).}
While SDBN improves robustness to a wide range of perturbations, its embedding-space $\ell_p$-ball constraint cannot capture \emph{tokenization-breaking} character edits, where a single change sends the resulting embedding far outside the continuous uncertainty region. To address this, \textsc{SDBN-h} augments standard FGSM perturbations with gradient-guided adversarial examples drawn from a discrete uncertainty set of single-character variants.

On \textsc{BLESS} with DeBERTa-v3 + LoRA trained on 1{,}000 clean samples,
SDBN-h yields the best robustness on tokenization-breaking types (see~\cref{tab:sdbn_h_results}),
improving by about +4–7\% while matching clean accuracy.
We include a qualitative robustness example illustrating tokenization-breaking character noise and SDBN-h behavior in Appendix~\ref{qualitive_example}.

\begin{table}[t]
\centering
\small
\caption{Generative task results under clean and noisy evaluation. \textbf{Top:} SQuAD exact match (EM) using LLaMA-3.2-1B with LoRA, trained on 200 samples. \textbf{Bottom:} TweetQA F1 using LLaMA-2-7B with LoRA, trained on 200 samples. SDBN-p uses generated adversarial variants during training. Best results in each column are shown in \textbf{bold}; second-best results are \underline{underlined}.}
\label{tab:generative}
\begin{tabular}{lcccc}
\toprule
\multicolumn{4}{c}{\textit{SQuAD (EM), LLaMA-3.2-1B}} \\
\midrule
Method & Clean & Swap-Word & Homophone \\
\midrule
Vanilla  & 58.92             & 32.44             & 47.28 \\
NEFTune  & \underline{59.72} & \underline{32.72} & \underline{48.08} \\
EDA      & 58.88             & 32.44             & 47.96 \\
FreeLB   & 57.00             & 31.64             & 44.04 \\
SMART    & 52.64             & 28.44             & 40.52 \\
SDBN-p   & \textbf{59.84}    & \textbf{35.08}    & \textbf{52.20} \\
\midrule
\multicolumn{4}{c}{\textit{TweetQA (F1), LLaMA-2-7B}} \\
\midrule
Method & Clean & Delete-Char & Delete-Word \\
\midrule
Vanilla  & 68.09              & 51.56              & 56.06 \\
NEFTune  & 69.34              & 55.13              & 54.30 \\
EDA      & 70.02  & 53.14              & 55.04 \\
FreeLB   & \underline{76.57}  & \underline{60.59}  & \underline{60.79} \\
SMART    & 66.71              & 49.54              & 51.13 \\
SDBN-p   & \textbf{80.81}     & \textbf{65.55}     & \textbf{64.15} \\
\bottomrule
\end{tabular}
\end{table}

\paragraph{Generative Tasks.}
Our initial continuous-perturbation variant was less effective on generative tasks (Appendix~\ref{generative_tasks_appendix}), motivating SDBN-p. We therefore evaluate on two generative benchmarks: SQuAD with LLaMA-3.2-1B and TweetQA with LLaMA-2-7B. Table~\ref{tab:generative} shows that SDBN-p improves performance over all baselines on both tasks, under both clean and noisy evaluation. Additional generative results and details are provided in Appendix~\ref{generative_tasks_appendix}.

\FloatBarrier

\begin{table}[h]
\centering
\small
\caption{\textbf{SDBN is substantially more effective with PEFT than with full fine-tuning.} The table reports the absolute accuracy gain (percentage points) achieved by adding SDBN to each training method on a low-resource subset of Banking77 (using DeBERTa-v3). For example, while SDBN improves LoRA by 23.6(\%) on clean data, it only improves full fine-tuning by 1.3(\%). This demonstrates that the adversarial signal is significantly more impactful within the constrained parameter subspaces of PEFT methods.}
\label{tab:peft_vs_fullft}
\begin{tabular}{lccccc}
\toprule
\textbf{Method} & \textbf{Clean} & \textbf{Replace} & \textbf{Delete} & \textbf{Swap} & \textbf{Avg gain} \\
\midrule
LoRA     & +23.6 & +18.8 & +18.7 & +17.1 & +19.6 \\
BitFit   & +16.0 & +11.2 & +12.8 & +11.3 & +12.8 \\
Adapter  & +13.3 & +9.4  & +9.8  & +6.2  & +9.7  \\
Full FT  & +1.3  & +0.9  & +0.5  & +0.8  & +0.9  \\
\bottomrule
\end{tabular}
\end{table}

\paragraph{PEFT vs.\ full fine-tuning under adversarial training.}
\label{peft_vs_full_ft}
We find that adversarial training is more effective when combined with PEFT than with full fine-tuning in low-resource settings. Table~\ref{tab:peft_vs_fullft} compares the gain of SDBN over the corresponding vanilla regime on a small subset of \textsc{Banking77} with DeBERTa-v3, evaluated on clean data and under three word-level corruptions. The gains are consistently substantial for PEFT methods, but negligible for full fine-tuning. This asymmetry suggests that PEFT's constrained parameter space may make the adversarial signal more focused and effective when data is scarce and noisy, whereas full fine-tuning is more vulnerable to overfitting adversarial examples in its much larger parameter number.

\FloatBarrier

\section{Conclusion}
We present SDBN (Small Data Big Noise), integrating adversarial training with PEFT to improve robustness in low-resource settings. Experiments show significant gains in robustness and accuracy across datasets and PEFT variants. We further extend the framework with two complementary strategies for discrete uncertainty set construction: SDBN-h, which incorporates gradient-guided character-level perturbations to address token-breaking typos, and SDBN-p, which leverages LLM-generated adversarial variants for generative tasks. The framework preserves PEFT's parameter efficiency while improving robustness to word-level and character-level noise and to domain shift, without requiring additional training data.


\newpage
\section{Limitations}
While our method substantially improves robustness to both word-level and character-level perturbations, several limitations remain. Generating adversarial embeddings requires an additional forward–backward pass per mini-batch. Although this overhead does not increase GPU memory usage, it does raise per-batch runtime and may become a bottleneck when scaling to very large models or training corpora. Moreover, the choice of perturbation radius $\epsilon$ remains sensitive across datasets. While our analysis provides a principled default, more adaptive or automated tuning strategies could further enhance robustness and ease adoption.
\bibliography{custom}

@article{lora,
  author    = {Edward Hu and Yelong Shen and Phillip Wallis and Zeyuan Allen-Zhu and Yuanzhi Li and Shean Wang and Lu Wang and Weizhu Chen},
  title     = {{LoRA}: Low-Rank Adaptation of Large Language Models},
  journal   = {arXiv preprint arXiv:2106.09685},
  year      = {2021},
  url       = {https://arxiv.org/abs/2106.09685}
}

@inproceedings{melora,
    title = "{MEL}o{RA}: {M}ini-{E}nsemble {L}ow-{R}ank {A}dapters for {P}arameter-{E}fficient {F}ine-{T}uning",
    author = "Ren, Pengjie and Shi, Chengshun and Wu, Shiguang and Zhang, Mengqi and Ren, Zhaochun and de Rijke, Maarten and Chen, Zhumin and Pei, Jiahuan",
    booktitle = "Proceedings of the 62nd Annual Meeting of the Association for Computational Linguistics (Volume 1: Long Papers)",
    month = aug,
    year = "2024",
    address = "Bangkok, Thailand",
    publisher = "Association for Computational Linguistics",
    url = "https://aclanthology.org/2024.acl-long.168/",
    doi = "10.18653/v1/2024.acl-long.168",
    pages = "3052--3064",
}

@inproceedings{adalora,
  author    = {Qingru Zhang and Minshuo Chen and Alexander Bukharin and Nikos Karampatziakis and Pengcheng He and Yu Cheng and Weizhu Chen and Tuo Zhao},
  title     = {{AdaLoRA}: {A}daptive {B}udget {A}llocation for {P}arameter-{E}fficient {F}ine-{T}uning},
  booktitle = {International Conference on Learning Representations (ICLR)},
  year      = {2023},
  url       = {https://arxiv.org/abs/2303.10512}
}

@article{qlora,
  author    = {Tim Dettmers and Artidoro Pagnoni and Ari Holtzman and Luke Zettlemoyer},
  title     = {{QLORA}: {E}fficient {F}inetuning of {Q}uantized {LLM}s},
  journal   = {arXiv preprint arXiv:2305.14314},
  year      = {2023},
  url       = {https://arxiv.org/abs/2305.14314}
}

@article{advlora,
  author    = {Yuheng Ji and Yue Liu and Zhicheng Zhang and Zhao Zhang and Yuting Zhao and Gang Zhou and Xingwei Zhang and Xinwang Liu and Xiaolong Zheng},
  title     = {{AdvLoRA}: {A}dversarial {L}ow-{R}ank {A}daptation of {V}ision-{L}anguage {M}odels},
  journal   = {arXiv preprint arXiv:2404.13425},
  year      = {2024},
  url       = {https://arxiv.org/abs/2404.13425}
}

@inproceedings{loft,
  author={Fu, Jiadong and Fang, Jiang and Sun, Jiyan and Zhuang, Shangyuan and Geng, Liru and Liu, Yinlong},
  booktitle={2024 International Joint Conference on Neural Networks (IJCNN)}, 
  title={{LoFT}: {LoRA}-{B}ased {E}fficient and {R}obust {F}ine-{T}uning {F}ramework for {A}dversarial {T}raining}, 
  year={2024},
  pages={1-8},
  doi={10.1109/IJCNN60899.2024.10651480}
}

@inproceedings{freelb,
  author    = {Chen Zhu and Yu Cheng and Zhe Gan and Siqi Sun and Tom Goldstein and Jingjing Liu},
  title     = {{FREELB}: {E}nhanced {A}dversarial {T}raining for {N}atural {L}anguage {U}nderstanding},
  booktitle = {International Conference on Learning Representations (ICLR)},
  year      = {2020},
  url       = {https://arxiv.org/abs/1909.11764v5}
}

@article{SHAHAM2018195,
  title = {Understanding adversarial training: {I}ncreasing local stability of supervised models through robust optimization},
  journal = {Neurocomputing},
  volume = {307},
  pages = {195-204},
  year = {2018},
  issn = {0925-2312},
  doi = {https://doi.org/10.1016/j.neucom.2018.04.027},
  url = {https://www.sciencedirect.com/science/article/pii/S0925231218304557},
  author = {Uri Shaham and Yutaro Yamada and Sahand Negahban},
}

@inproceedings{bitfit,
  title={{BitFit}: {S}imple {P}arameter-efficient {F}ine-tuning for {T}ransformer-based {M}asked {L}anguage-models},
  author={Ben-Zaken, Elad and Ravfogel, Shauli and Goldberg, Yoav},
  booktitle={Proceedings of the 62nd Annual Meeting of the Association for Computational Linguistics (ACL)},
  year={2021},
  organization={Association for Computational Linguistics}
}

@inproceedings{adapter,
  title={{P}arameter-{E}fficient {T}ransfer {L}earning for {NLP}},
  author={Houlsby, Neil and Giurgiu, Andrei and Jastrzebski, Stanisław and Morrone, Bruna and de Laroussilhe, Quentin and Gesmundo, Andrea and Attariyan, Mona and Gelly, Sylvain},
  booktitle={Proceedings of the 36th International Conference on Machine Learning (ICML)},
  volume={97},
  pages={2790--2799},
  year={2019},
  organization={PMLR}
}

@inproceedings{TowardsRobustAndGeneralizedPEFT,
  title={Towards {R}obust and {G}eneralized {P}arameter-{E}fficient {F}ine-{T}uning for {N}oisy {L}abel {L}earning},
  author={Kim, Yeachan and Kim, Junho and Lee, SangKeun},
  booktitle={Proceedings of the 62nd Annual Meeting of the Association for Computational Linguistics (ACL)},
  pages={5922--5936},
  year={2024},
  organization={Association for Computational Linguistics}
}

@article{madry2019deep,
  title={Towards deep learning models resistant to adversarial attacks},
  author={Madry, Aleksander and Makelov, Aleksandar and Schmidt, Ludwig and Tsipras, Dimitris and Vladu, Adrian},
  journal={arXiv preprint arXiv:1706.06083},
  year={2019},
   url={https://arxiv.org/abs/1706.06083}
}

@inproceedings{goodfellow-etal-2015-iclr,
  title     = {Explaining and Harnessing Adversarial Examples},
  author    = {Goodfellow, Ian J. and Shlens, Jonathon and Szegedy, Christian},
  booktitle = {International Conference on Learning Representations (ICLR)},
  year      = {2015},
  address   = {San Diego, CA, USA},
  url       = {https://arxiv.org/abs/1412.6572},
}

@book{ben2009robust,
  title={Robust Optimization},
author = {Ben-Tal, A. and Nemirovski, Arkadi and El Ghaoui, Laurent},
publisher = {Princeton University Press},
  address   = {Princeton, NJ},
  volume={2},
  pages={35--53},
  year={2009}
}

@article{banking77,
  author    = {Iñigo Casanueva and Tadas Temčinas and Daniela Gerz and Matthew Henderson and Ivan Vulić},
  title     = {Efficient Intent Detection with Dual Sentence Encoders},
  journal   = {arXiv preprint arXiv:2003.04807},
  year      = {2020},
  url       = {http://arxiv.org/abs/2003.04807}
}

@inproceedings{trec,
  author    = {Amit Singhal and Steve Abney and Michiel Bacchiani and Michael Collins and Donald Hindle and Fernando Pereira},
  title     = {{AT\&T} at {TREC}-8},
  booktitle = {Proceedings of the Eighth Text REtrieval Conference (TREC-8)},
  year      = {1999},
  organization = {NIST},
  url       = {https://trec.nist.gov/pubs/trec8/papers/att8.qa.pdf}
}

@inproceedings{news20,
  author = {Ken Lang},
  title = {Newsweeder: {L}earning to filter netnews},
  year = {1995},
  booktitle = {Proceedings of the Twelfth International Conference on Machine Learning},
  pages = {331-339}
}

@book{boyd2004convex,
  title={Convex optimization},
  author={Boyd, Stephen and Vandenberghe, Lieven},
  publisher={Cambridge University Press},
  year={2004}
}

@inproceedings{neftune,
  title={NEFTune: Noisy Embeddings Improve Instruction Finetuning},
  author={Jain, Neel and Chiang, Ping-yeh and Wen, Yuxin and Kirchenbauer, John and Chu, Hong-Min and Somepalli, Gowthami and Bartoldson, Brian R. and Kailkhura, Bhavya and Schwarzschild, Avi and Saha, Aniruddha and Goldblum, Micah and Geiping, Jonas and Goldstein, Tom},
  booktitle={Preprint},
  year={2023},
  url={https://arxiv.org/abs/2310.05914}
}

@article{prompt,
  title     = {P-Tuning v2: Prompt Tuning Can Be Comparable to Fine-tuning Universally Across Scales and Tasks},
  author    = {Xiao Liu and Kaixuan Ji and Yicheng Fu and Weng Lam Tam and Zhengxiao Du and Zhilin Yang and Jie Tang},
  journal   = {arXiv preprint arXiv:2110.07602},
  year      = {2021},
  url       = {https://arxiv.org/abs/2110.07602},
  note      = {Version 3}
}

@article{prefix,
  title     = {Prefix-Tuning: Optimizing Continuous Prompts for Generation},
  author    = {Li, Xiang Lisa and Liang, Percy},
  journal   = {arXiv preprint arXiv:2101.00190},
  year      = {2021},
  url       = {https://arxiv.org/abs/2101.00190}
}

@article{smart,
  title={{SMART}: Robust and Efficient Fine‑Tuning for Pre‑trained Natural Language Models through Principled Regularized Optimization},
  author={Jiang, Haoming and He, Pengcheng and Chen, Weizhu and Liu, Xiaodong and Gao, Jianfeng and Zhao, Tuo},
  journal={arXiv preprint arXiv:1911.03437v5},
  year={2019},
  url={https://arxiv.org/abs/1911.03437v5}
}

@article{miyato2016adversarial,
  title={{Adversarial Training Methods for Semi-Supervised Text Classification}},
  author={Miyato, Takeru and Dai, Andrew M and Goodfellow, Ian},
  journal={arXiv preprint arXiv:1605.07725v4},
  year={2021},
  url={https://arxiv.org/abs/1605.07725v4}
}

@inproceedings{DAE,
  title     = {Extracting and Composing Robust Features with Denoising Autoencoders},
  author    = {Vincent, Pascal and Larochelle, Hugo and Bengio, Yoshua and Manzagol, Pierre-Antoine},
  booktitle = {Proceedings of the 25th International Conference on Machine Learning (ICML 2008)},
  pages     = {1096--1103},
  year      = {2008},
  url       = {https://dl.acm.org/doi/10.1145/1390156.1390294}
}

@inproceedings{Pu2023Empirical,
  title     = {Empirical Analysis of the Strengths and Weaknesses of PEFT Techniques for LLMs},
  author    = {Pu, George and Jain, Anirudh and Yin, Jihan and Kaplan, Russell},
  booktitle = {Proceedings of the Workshop on Understanding Foundation Models at ICLR},
  year      = {2023},
  url       = {https://arxiv.org/abs/2304.14999}
}

@inproceedings{abufarha-etal-2021-arsarcasm-v2,
title = "Overview of the WANLP 2021 Shared Task on Sarcasm and Sentiment Detection in Arabic",
    author = "Abu Farha, Ibrahim  and
    Zaghouani, Wajdi  and
    Magdy, Walid",
    booktitle = "Proceedings of the Sixth Arabic Natural Language Processing Workshop",
    month = {April},
    year = "2021",
    }

@inproceedings{Devlin2019BERT,
  title     = {BERT: Pre-training of Deep Bidirectional Transformers for Language Understanding},
  author    = {Devlin, Jacob and Chang, Ming-Wei and Lee, Kenton and Toutanova, Kristina},
  booktitle = {Proceedings of the 2019 Conference of the North American Chapter of the Association for Computational Linguistics: Human Language Technologies, Volume 1 (Long and Short Papers)},
  pages     = {4171--4186},
  year      = {2019}
}

@inproceedings{He2023DeBERTaV3,
  title     = {{DeBERTaV3}: Improving {DeBERTa} using ELECTRA-Style Pre-Training with Gradient-Disentangled Embedding Sharing},
  author    = {He, Pengcheng and Gao, Jianfeng and Chen, Weizhu},
  booktitle = {Proceedings of the International Conference on Learning Representations (ICLR)},
  year      = {2023}
}

@article{squad,
  author    = {Pranav Rajpurkar and Jian Zhang and Konstantin Lopyrev and Percy Liang},
  title     = {{SQuAD}: 100,000+ Questions for Machine Comprehension of Text},
  journal   = {arXiv preprint arXiv:1606.05250},
  year      = {2016},
  url       = {https://arxiv.org/abs/1606.05250}
}

@inproceedings{bless,
  author    = {Marco Baroni and Alessandro Lenci},
  title     = {How we {BLESS}ed distributional semantic evaluation},
  booktitle = {Proceedings of the GEMS 2011 Workshop on Geometrical Models of Natural Language Semantics, EMNLP 2011},
  pages     = {1--10},
  year      = {2011},
  address   = {Edinburgh, Scotland, UK},
  publisher = {Association for Computational Linguistics},
  url       = {http://clic.cimec.unitn.it/distsem}
}

@inproceedings{wildnlp,
  title={Models in the Wild: On Corruption Robustness of Neural NLP Systems},
  author={Rychalska, Barbara and Basaj, Dominika and Gosiewska, Alicja and Biecek, Przemys{\l}aw},
  booktitle={Neural Information Processing (ICONIP 2019)},
  series={Lecture Notes in Computer Science},
  volume={11955},
  pages={235--247},
  publisher={Springer},
  year={2019},
  url={https://link.springer.com/chapter/10.1007/978-3-030-36718-3_20},
  doi={10.1007/978-3-030-36718-3_20}
}

@inproceedings{hotflip,
  title={{HotFlip}: White-Box Adversarial Examples for Text Classification},
  author={Ebrahimi, Javid and Rao, Anyi and Lowd, Daniel and Dou, Dejing},
  booktitle={Proceedings of the 56th Annual Meeting of the Association for Computational Linguistics (ACL)},
  year={2018},
  url={https://aclanthology.org/P18-2006}
}

@inproceedings{dialectnoise,
  title={Improving Zero-Shot Cross-lingual Transfer Between Closely Related Languages by Injecting Character-Level Noise},
  author={Aepli, No{\"e}mi and Sennrich, Rico},
  booktitle={Findings of the Association for Computational Linguistics: ACL 2022},
  year={2022},
  pages={4074--4083},
  url={https://aclanthology.org/2022.findings-acl.321/}
}

@article{eda,
  author    = {Jason Wei and Kai Zou},
  title     = {{EDA}: {E}asy {D}ata {A}ugmentation {T}echniques for {B}oosting {P}erformance on {T}ext {C}lassification {T}asks},
  journal   = {arXiv preprint arXiv:1901.11196},
  year      = {2019},
  url       = {https://arxiv.org/abs/1901.11196}
}

@misc{openai_gpt52_systemcard,
  title        = {Update to {GPT-5} System Card: {GPT-5.2}},
  author       = {{OpenAI}},
  year         = {2025},
  month        = dec,
  howpublished = {Technical report (PDF)},
  url          = {https://cdn.openai.com/pdf/3a4153c8-c748-4b71-8e31-aecbde944f8d/oai_5_2_system-card.pdf}
}

@article{llama3herd,
  author    = {{Llama Team}},
  title     = {The {Llama} 3 Herd of Models},
  journal   = {arXiv preprint arXiv:2407.21783},
  year      = {2024},
  url       = {https://arxiv.org/abs/2407.21783}
}

@article{zeng-etal-2023-certified,
  author    = {Jiehang Zeng and Jianhan Xu and Xiaoqing Zheng and Xuanjing Huang},
  title     = {Certified Robustness to Text Adversarial Attacks by Randomized [MASK]},
  journal   = {Computational Linguistics},
  volume    = {49},
  number    = {2},
  pages     = {395--429},
  year      = {2023},
  url       = {https://aclanthology.org/2023.cl-2.5.pdf}
}

@inproceedings{huang-etal-2019-achieving,
  author    = {Po-Sen Huang and Robert Stanforth and Johannes Welbl and Chris Dyer and Dani Yogatama and Sven Gowal and Krishnamurthy Dvijotham and Pushmeet Kohli},
  title     = {Achieving Verified Robustness to Symbol Substitutions via Interval Bound Propagation},
  booktitle = {Proceedings of the 2019 Conference on Empirical Methods in Natural Language Processing and the 9th International Joint Conference on Natural Language Processing ({EMNLP-IJCNLP})},
  pages     = {4083--4093},
  year      = {2019},
  url       = {https://aclanthology.org/D19-1419.pdf}
}

@inproceedings{jia-etal-2019-certified,
  author    = {Robin Jia and Aditi Raghunathan and Kerem Gokhan G{\"u}lcehre and Percy Liang},
  title     = {Certified Robustness to Adversarial Word Substitutions},
  booktitle = {Proceedings of the 2019 Conference on Empirical Methods in Natural Language Processing and the 9th International Joint Conference on Natural Language Processing ({EMNLP-IJCNLP})},
  pages     = {4129--4142},
  year      = {2019},
  url       = {https://aclanthology.org/D19-1423.pdf}
}

@inproceedings{zhou-etal-2021-defense,
  author    = {Yi Zhou and Xiaoqing Zheng and Cho-Jui Hsieh and Kai-Wei Chang and Xuanjing Huang},
  title     = {Defense against Synonym Substitution-based Adversarial Attacks via Dirichlet Neighborhood Ensemble},
  booktitle = {Proceedings of the 59th Annual Meeting of the Association for Computational Linguistics and the 11th International Joint Conference on Natural Language Processing (Volume 1: Long Papers)},
  pages     = {5482--5492},
  year      = {2021},
  url       = {https://aclanthology.org/2021.acl-long.426.pdf}
}

@inproceedings{ivgi-berant-2021-achieving,
  author    = {Maor Ivgi and Jonathan Berant},
  title     = {Achieving Model Robustness through Discrete Adversarial Training},
  booktitle = {Proceedings of the 2021 Conference on Empirical Methods in Natural Language Processing},
  pages     = {1529--1544},
  year      = {2021},
  url       = {https://aclanthology.org/2021.emnlp-main.115.pdf}
}

@mastersthesis{rusert2025investigating,
  author = {Ajinkya More},
  title = {Investigating the Robustness of Parameter Efficient Fine Tuning Methods Against Adversarial Attacks in Natural Language Processing},
  school = {Purdue University},
  address = {Fort Wayne, Indiana},
  year = {2025},
  month = aug
}

@inproceedings{xiong-etal-2019-tweetqa,
  title     = {{TWEETQA}: A Social Media Focused Question Answering Dataset},
  author    = {Xiong, Wenhan and Wu, Jiawei and Wang, Hong and Kulkarni, Vivek and Yu, Mo and Chang, Shiyu and Guo, Xiaoxiao and Wang, William Yang},
  booktitle = {Proceedings of the 57th Annual Meeting of the Association for Computational Linguistics},
  pages     = {5020--5031},
  year      = {2019},
  address   = {Florence, Italy},
  publisher = {Association for Computational Linguistics}
}

@article{qwen2025qwen25,
  title         = {Qwen2.5 Technical Report},
  author        = {{Qwen Team}},
  journal       = {arXiv preprint arXiv:2412.15115},
  year          = {2025},
  eprint        = {2412.15115},
  archivePrefix = {arXiv},
  primaryClass  = {cs.CL}
}

@article{touvron2023llama2,
  title         = {Llama 2: Open Foundation and Fine-Tuned Chat Models},
  author        = {Touvron, Hugo and Martin, Louis and Stone, Kevin and Albert, Peter and Almahairi, Amjad and Babaei, Yasmine and Bashlykov, Nikolay and Batra, Soumya and Bhargava, Prajjwal and Bhosale, Shruti and Bikel, Dan and Blecher, Lukas and Canton Ferrer, Cristian and Chen, Moya and Cucurull, Guillem and Esiobu, David and Fernandes, Jude and Fu, Jeremy and Fu, Wenyin and Fuller, Brian and Gao, Cynthia and Goswami, Vedanuj and Goyal, Naman and Hartshorn, Anthony and Hosseini, Saghar and Hou, Rui and Inan, Hakan and Kardas, Marcin and Kerkez, Viktor and Khabsa, Madian and Kloumann, Isabel and Korenev, Artem and Koura, Punit Singh and Lachaux, Marie-Anne and Lavril, Thibaut and Lee, Jenya and Liskovich, Diana and Lu, Yinghai and Mao, Yuning and Martinet, Xavier and Mihaylov, Todor and Mishra, Pushkar and Molybog, Igor and Nie, Yixin and Poulton, Andrew and Reizenstein, Jeremy and Rungta, Rashi and Saladi, Kalyan and Schelten, Alan and Silva, Ruan and Smith, Eric Michael and Subramanian, Ranjan and Tan, Xiaoqing Ellen and Tang, Binh and Taylor, Ross and Williams, Adina and Xiang, Jian and Xu, Puxin and Yan, Zheng and Zarov, Iliyan and Zhang, Yuchen and Fan, Angela and Kambadur, Melanie and Narang, Sharan and Rodriguez, Aurelien and Stojnic, Robert and Edunov, Sergey and Scialom, Thomas},
  journal       = {arXiv preprint arXiv:2307.09288},
  year          = {2023},
  eprint        = {2307.09288},
  archivePrefix = {arXiv},
  primaryClass  = {cs.CL}
}

\newpage
\appendix

\section{Additional Preliminaries}
\subsection{Low-Rank Adaptation (LoRA)}
LoRA \cite{lora} is a parameter-efficient fine-tuning method that adapts large-scale pre-trained models by learning low-rank updates. Consider a weight matrix $W_0 \in \mathbb{R}^{d \times k}$ in a pre-trained model. Instead of directly fine-tuning $W_0$, LoRA modifies it as:
\[
W = W_0 + \Delta W,\quad 
\]
\[
\text{where } \Delta W = AB,
\]
with $A \in \mathbb{R}^{d \times r}$ and $B \in \mathbb{R}^{r \times k}$, and $r \ll \min(d,k)$. This low-rank factorization significantly reduces the number of trainable parameters while retaining performance.

\section{Geometry of \texorpdfstring{$p$}{p}-Norm Uncertainty Sets and Perturbation Behavior}
\label{app:norms}
Let $g = \nabla \mathcal{L}_{\theta,y}(x)$.
Different choices of norm \(p\) determine distinct perturbation characteristics. The optimal perturbation \(\delta^*\) can be approximated by a single steepest ascent step, yielding \(\tilde{\delta}\) that maximizes the inner product in \cref{eq:optimal_delta}. Steepest ascent determines the direction of maximal increase by optimizing the inner product with the function's gradient, subject to the given step size and the specific choice of norm, thereby defining \(\tilde{\delta}\). Choosing \(\tilde{\delta}\) from \(\ell_{\infty}\) ball, generates perturbation where each entry of $x$ is modified by the same amount in the direction determined by the sign of the gradient $g$, (\(\tilde{\delta}=\epsilon \cdot \operatorname{sign}(g)\)) making it particularly suitable for attacks on vision models where imperceptible changes are crucial - every pixel is modified by a small $\epsilon$. A practical adversarial training approach leveraging the \(\ell_{\infty}\) uncertainty set is the Fast Gradient Sign Method (FGSM) \citep{goodfellow-etal-2015-iclr} which defines \(\tilde{\delta}\) in a similar way.
For \(\ell_2\) ball, the steepest ascent produces perturbations aligned with the direction of $g$, and choosing \(\tilde{\delta}\) from \(\ell_1\) ball yields sparse perturbations where only a few entries corresponding to the components of $g$ with largest absolute values are modified.
\section{Addition method}


\makeatletter
\algrenewcommand\alglinenumber[1]{\footnotesize \arabic{ALG@line}:}
\makeatother
\algrenewcommand\algorithmicrequire{\textbf{Require:}}

\begin{algorithm}[ht]
  \caption{\textbf{SDBN} -- Small Data Big Noise:\\ Adversarial Training with $\ell_\infty$}
  \label{adv_training_l_inf}
  \begin{algorithmic}[1]
    \Require Input batch $X$, labels $Y$, embedding extractor $E$, model $f(\cdot;\theta)$, initial $\epsilon$
    \State $\mathbf{e} \gets E(X)$
    \State $\mathcal{L}_{\text{clean}} \gets \mathcal{L}(f_\theta(\mathbf{e}),\,Y)$
    \State $\mathbf{g} \gets \nabla_{\mathbf{e}}\,\mathcal{L}_{\text{clean}}$
    \State $\delta \gets \epsilon\,\operatorname{sign}(\mathbf{g})$
    \State $\mathbf{e}_{\text{adv}} \gets \mathbf{e} + \delta$
    \State $\mathcal{L}_{\text{adv}} \gets \mathcal{L}(f_\theta(\mathbf{e}_{\text{adv}}),\,Y)$
    \State \textbf{Update} $\theta$ using $\mathcal{L}_{\text{adv}}$
  \end{algorithmic}
\end{algorithm}

\subsection{Epsilon Selection.}
\label{epsilon_slection}
To maintain \(\mathcal{L}_{\text{adv}}\) consistently higher than \(\mathcal{L}_{\text{clean}}\), we empirically determine an appropriately small \(\epsilon\) value (see \cref{eps_exp_apendix}). Our implementation includes an adaptive mechanism with a parameter $K$ that adjusts \(\epsilon\): when \(\mathcal{L}_{\text{adv}}\) falls below \(\mathcal{L}_{\text{clean}}\), we conduct a line-search \citep{boyd2004convex} based search while iteratively reducing \(\epsilon\) by a factor of 10 for up to $K$ iterations until the constraint \(\mathcal{L}_{\text{adv}} > \mathcal{L}_{\text{clean}}\) is satisfied. In practice, by choosing the appropriate \(\epsilon\), the constraint is satisfied using just one iteration (see \cref{eps_exp_apendix}).

\subsection{SDBN-h}
\label{sdbn-h_appendix}
During training, each mini-batch is split into two parts: we apply SDBN (embedding-space $\ell_\infty$ perturbations) to the first part and SDBN-h (character-variant selection from $\mathcal{C}(x)$) to the remaining part (see \Cref{alg:sdbn_h_single}).

\begin{algorithm}[H]
\caption{\textsc{SDBN}-h: Training Step}
\label{alg:sdbn_h_single}
\begin{algorithmic}[1]
\Require Clean input $x$, label $y$, model $f(\cdot;\theta)$, loss $\mathcal{L}$, embedding $E(\cdot)$
\Require Noise-function set $\mathcal{H}$, where each $h \in \mathcal{H}$ maps $(x,i)\mapsto z$ by applying a character-level edit to $x$ at index $i$
\State $g \leftarrow \nabla_{E(x)}\,\mathcal{L}\!\left(f(E(x);\theta),\, y\right)$ \Comment{gradient of the clean example}
\State Sample $h \sim \textsc{Uniform}(\mathcal{H})$
\State $\mathcal{C}(x) \leftarrow \{\, z_i = h(x,i)\ :\ i=1,\ldots,|x| \,\}$ \Comment{apply $h$ at all character indices}
\State $z^\star \leftarrow \arg\max_{z \in \mathcal{C}(x)} \left\langle g,\ E(z)-E(x)\ \right\rangle$
\State \textbf{Update} $\theta$ using $\mathcal{L}\!\left(f(E(z^\star);\theta),\, y\right)$
\end{algorithmic}
\end{algorithm}

\paragraph{Character perturbation types.}
The discrete uncertainty set $\mathcal{C}(x)$ consists of single-character variants generated by the following operations:

\begin{itemize}[leftmargin=*, itemsep=2pt]
    \item \textbf{Delete character:} Remove one character (e.g., ``card'' $\rightarrow$ ``crd'')
    \item \textbf{Swap characters:} Swap two adjacent characters (e.g., ``card'' $\rightarrow$ ``acrd'')
    \item \textbf{Double character:} Duplicate one character (e.g., ``card'' $\rightarrow$ ``carrd'')
    \item \textbf{Phonetic replacement:} Replace with phonetically similar character (e.g., ``phone'' $\rightarrow$ ``fone'')
    \item \textbf{Insert character:} Insert a random character (e.g., ``card'' $\rightarrow$ ``ca1rd'')
    \item \textbf{Cyrillic substitution:} Replace with visually similar Cyrillic character (e.g., ``card'' $\rightarrow$ ``ca{{\foreignlanguage{russian}{я}}}d'')
    \item \textbf{Random capitalization:} Change case of one character (e.g., ``card'' $\rightarrow$ ``caRd'')
\end{itemize}
For each input $x$, one perturbation type is randomly selected, and all single-edit variants under that type form $\mathcal{S}(x)$.

\subsection{SDBN-p}
\label{sdbn-p_appendix}
\begin{algorithm}[H]
\caption{SDBN-p - Training Step}
\label{alg:sdbn_p_single}
\begin{algorithmic}[1]
\Require Input $x$, label $y$, model $f(\cdot;\theta)$, loss $\mathcal{L}$, embedding $E(\cdot)$
\Require Pre-computed variants $\mathcal{P}(x) = \{z_1, \dots, z_k\}$
\State $z^\star \leftarrow \arg\max_{z_i \in \mathcal{P}(x)} \mathcal{L}\!\left(f(E(z_i);\theta),\, y\right)$ \Comment{select variant with maximum explicit loss}
\State $\mathcal{L}_{p} \leftarrow \mathcal{L}\!\left(f(E(z^\star);\theta),\, y\right)$
\State \textbf{Update} $\theta$ using $\mathcal{L}_{p}$
\end{algorithmic}
\end{algorithm}

\paragraph{LLM-generated adversarial variants.}\ The discrete uncertainty set $\mathcal{P}(x)$ is pre-computed offline by prompting an LLM (e.g., GPT-5.2) to generate $k$ adversarial variants for each training example. The variants introduce noise while preserving the semantic meaning and expected output, aiming to improve general robustness[cite: 1348]. This generation is performed once before training begins. During training, at each epoch, we explicitly compute the loss for each variant and select the one that maximizes the training objective. Because these variants can involve structural changes that fall outside the local linear region of the clean input, this explicit selection ensures we identify the true worst-case semantic neighbor. Importantly, since model parameters evolve during training, the worst-case variant may change from epoch to epoch, allowing the model to be exposed to different challenging examples throughout optimization.

\paragraph{Example prompt and output. (GPT-5.2)}
\begin{quote}
\small
\textbf{Prompt to LLM:} Generate 5 adversarial variants of the following input. Each variant should preserve the meaning and expected output, but challenge the model by increasing the loss size (cross-entropy), so that it will achieve robustness to noise.

\textbf{Input:} ``Context: Because of its Catholic identity, a number of religious buildings stand on campus. [...] The Grotto of Our Lady of Lourdes, which was built in 1896, is a replica of the original in Lourdes, France. [...] Question: In what year was the Grotto of Our Lady of Lourdes at Notre Dame constructed? Answer:''

\textbf{Expected output:} ``1896''

\textbf{Generated variant (1 of 5):} ``Context: Because of its Catholic identity, a number of religious buildings stand on camtpus. The Old College building has become oneofMtwo seminaries [...] The Grotto of Our Lady of Lourdes, which was built in 1896, is a replica of the original in Lourdes, France. [...] Question: In what yea waIs the Grotto of Our Lady of Lourdes at Notre Dame constructed? Answer:''
\end{quote}
The LLM introduces realistic character-level corruptions (e.g., ``campus'' $\rightarrow$ ``camtpus'', ``year was'' $\rightarrow$ ``yea waIs'') that break tokenization while preserving answerability.

\subsection{Algorithm $\ell_{2}$}
\label{pseudo_code_l2}

\begin{algorithm}[H]
\caption{Adversarial Training with $\ell_{2}$}
\label{alg:adv_training_l2}
\begin{algorithmic}[1]
\Require Input batch $X$, labels $Y$, embedding extractor $E$, model $f(\cdot;\theta)$, initial $\epsilon$
\State $e \leftarrow E(X)$
\State $\mathcal{L}_{\text{clean}} \leftarrow \mathcal{L}(f_\theta(e), Y)$
\State $g \leftarrow \nabla_{e}\,\mathcal{L}_{\text{clean}}$
\State $\delta \leftarrow \epsilon \cdot \dfrac{g}{\|g\|_{2}}$
\State $e_{\text{adv}} \leftarrow e + \delta$
\State $\mathcal{L}_{\text{adv}} \leftarrow \mathcal{L}(f_\theta(e_{\text{adv}}), Y)$
\State \textbf{Update} $\theta$ using $\mathcal{L}_{\text{adv}}$
\end{algorithmic}
\end{algorithm}

\subsection{Algorithm $\ell_{1}$}
\label{pseudo_code_l1}

\begin{algorithm}[H]
\caption{Adversarial Training with $\ell_{1}$}
\label{alg:adv_training_l1}
\begin{algorithmic}[1]
\Require Input batch $X$, labels $Y$, embedding extractor $E$, model $f(\cdot;\theta)$, initial $\epsilon$
\State $e \leftarrow E(X)$
\State $\mathcal{L}_{\text{clean}} \leftarrow \mathcal{L}(f_\theta(e), Y)$
\State $g \leftarrow \nabla_{e}\,\mathcal{L}_{\text{clean}}$
\State $i^\star \leftarrow \arg\max_{i} |g_i|$
\State $\delta \leftarrow 0$ \Comment{same shape as $e$}
\State $\delta_{i^\star} \leftarrow \epsilon \cdot \operatorname{sign}(g_{i^\star})$ \Comment{add perturbation only at the max-magnitude entry}
\State $e_{\text{adv}} \leftarrow e + \delta$
\State $\mathcal{L}_{\text{adv}} \leftarrow \mathcal{L}(f_\theta(e_{\text{adv}}), Y)$
\State \textbf{Update} $\theta$ using $\mathcal{L}_{\text{adv}}$
\end{algorithmic}
\end{algorithm}

\FloatBarrier

\subsection{Evaluate Epsilon and Norms} \label{norms_method_appendix} \label{eps_exp_apendix}
\begin{table}[t]
  \centering
  \tiny
  \footnotesize
  \caption{Evaluation of different $\ell_p$ norms for varying $\epsilon$ values on \textit{Banking77} (1,000 samples). Based on these results, we chose $\ell_{\infty}$ with $\epsilon=10^{-4}$ as the best option.}
  \label{tab:lnorm_evaluation}

  \setlength{\tabcolsep}{4pt}
  \renewcommand{\arraystretch}{1.05}

  \begin{subtable}{\linewidth}
    \centering
    \begin{tabular}{@{}c|cccc@{}}
    
      \toprule
      $\ell_p\ /\ \epsilon$ & $10^{1}$ & $10^{0}$ & $10^{-1}$ & $10^{-2}$ \\
      \midrule
      $\ell_{\infty}$ & $2.26\!\pm\!0.71$ & $2.75\!\pm\!0.95$ & $2.80\!\pm\!1.30$ & $4.41\!\pm\!3.09$ \\
      $\ell_{1}$      & $65.58\!\pm\!1.54$ & $67.20\!\pm\!1.00$ & $68.13\!\pm\!0.96$ & $67.79\!\pm\!0.49$ \\
      $\ell_{2}$      & $9.38\!\pm\!11.16$ & $70.24\!\pm\!2.23$ & $70.18\!\pm\!0.85$ & $68.32\!\pm\!1.57$ \\
      \bottomrule
      
    \end{tabular}
    \caption*{(a) Larger $\epsilon$ values}
    
  \end{subtable}

  \vspace{0.35em}

  \begin{subtable}{\linewidth}
    \centering
    \begin{tabular}{@{}c|ccc@{}}
      \toprule
      $\ell_p\ /\ \epsilon$ & $10^{-3}$ & $10^{-4}$ & $10^{-5}$ \\
      \midrule
      $\ell_{\infty}$ & $63.03\!\pm\!5.72$ & $\mathbf{72.86\!\pm\!1.24}$ & $67.43\!\pm\!0.87$ \\
      $\ell_{1}$      & $67.09\!\pm\!1.60$ & $67.03\!\pm\!1.09$ & $67.62\!\pm\!1.14$ \\
      $\ell_{2}$      & $67.80\!\pm\!1.18$ & $66.88\!\pm\!1.77$ & $67.38\!\pm\!1.34$ \\
      \bottomrule
     
    \end{tabular}
    \caption*{(b) Smaller $\epsilon$ values}
  \end{subtable}

\end{table}

\label{injected_layer}
\begin{table}[H]
    \small
    \centering
    \caption{\textbf{Perturbation Layer Impact on Model Performance.} Classification accuracy comparison when applying adversarial noise at different layers in BERT-base with LoRA, trained on 1000 samples from Banking77. Perturbations at the embedding layer dramatically outperform those at any encoder layer, highlighting that input-level modifications more effectively capture realistic linguistic variations.}
    \label{tab:perturbation_location}
    \begin{tabular}{l|c}
        \toprule
        \textbf{Perturbed Layer} & 
        \textbf{Accuracy ($\%$)} \\
        \midrule
        Encoder 1 & 6.71 $\pm$ 2.18 \\
        Encoder 5 & 6.17 $\pm$ 1.73 \\
        Encoder 10 & 6.32 $\pm$ 1.95 \\
        \textbf{Embeddings} & \textbf{57.64 $\pm$ 5.04} \\
        \bottomrule
    \end{tabular}
\end{table}

\subsection{Justification for Using $\ell_{\infty}$ Norm with $\epsilon=10^{-4}$}
\label{intution}
\begin{figure*}[t]
    \centering
    \includegraphics[width=\textwidth,keepaspectratio]{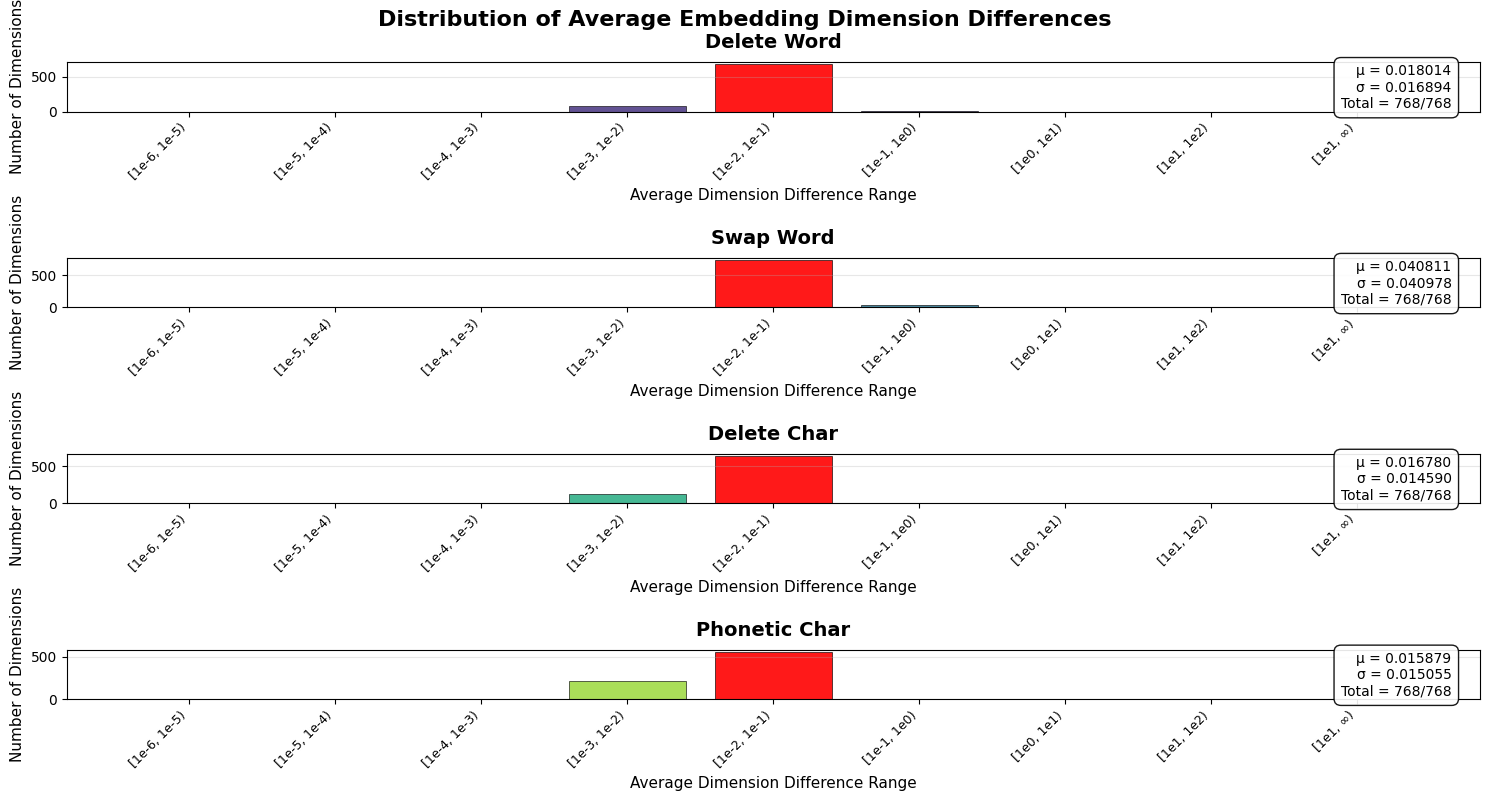}
    \caption{Histogram of embedding differences between noisy and clean sentences across multiple noise types. 
    The values cluster in a narrow, symmetric range around zero, indicating that realistic noise introduces 
    small but bounded shifts across embedding dimensions rather than sparse or extreme deviations.}
    \label{fig:histt}
\end{figure*}

\begin{figure*}[t]
  \centering
  \includegraphics[
    width=\textwidth,          
    height=0.42\textheight,    
    keepaspectratio,
    trim=6 4 6 4,clip
  ]{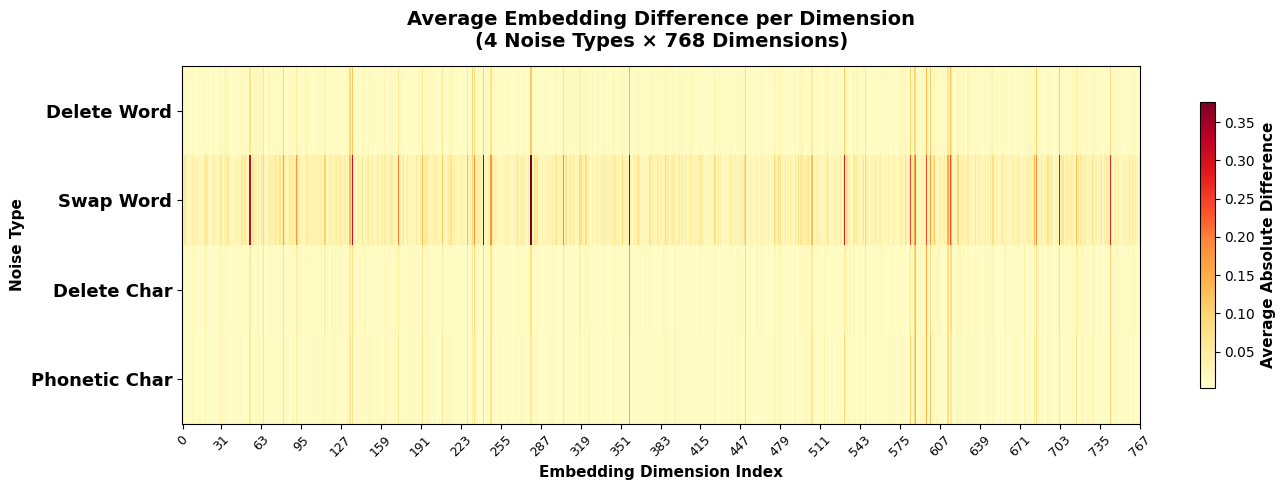}
  \caption{Heatmap of coordinate-wise embedding differences for multiple noise types.
  Each row is a noise type; each column is an embedding dimension. Similar magnitudes
  across rows indicate roughly uniform low-magnitude perturbations consistent with an
  $\ell_{\infty}$ ball.}
  \label{fig:heat_mp}
\end{figure*}

To justify our design choice, we analyzed how realistic textual noise affects sentence embeddings compared to clean data. Specifically, we took clean sentences and applied multiple noise types (e.g., word deletion, swap, replacement, case changes, character edits). For each noisy sentence, we computed the difference between its embedding vector and that of the corresponding clean sentence. We then visualized these embedding-level perturbations in two complementary ways: a histogram of coordinate-wise differences (Fig.~\ref{fig:histt}) and a heatmap of difference magnitudes across embedding indices and noise types (Fig.~\ref{fig:heat_mp}).

The histogram in Fig.~\ref{fig:histt} shows that the perturbation values are distributed in a narrow range, approximately symmetric around zero, with no heavy tails. This indicates that noise does not create sparse, extreme deviations, but rather small, bounded shifts across many embedding dimensions. Similarly, the heatmap in Fig.~\ref{fig:heat_mp} demonstrates that all noise types induce perturbations of comparable magnitude, uniformly spread across embedding coordinates. Together, these observations suggest that realistic textual noise behaves like a low-magnitude, approximately uniform distribution across embedding dimensions.

These empirical findings motivate the use of the $\ell_{\infty}$ norm for adversarial perturbations, since the $\ell_{\infty}$ ball constrains each coordinate to be shifted by the same amount, reflecting the bounded, coordinate-wise nature of realistic noise. From our norm comparison study (Table~\ref{tab:lnorm_evaluation}), we found that $\ell_{\infty}$ with $\epsilon=10^{-4}$ yielded the best trade-off: large enough to consistently increase the adversarial loss $L_{\text{adv}}$ over the clean loss $L_{\text{clean}}$, yet small enough to preserve semantic proximity to the clean embeddings. Thus, this choice is both theoretically justified by the geometry of observed perturbations and empirically validated by robustness improvements.

\FloatBarrier

\label{noise_types}

\begin{table*}[!b]
    \centering
    \footnotesize 
    \small
    \caption{Types of noise perturbations used for robustness evaluation}
    \label{tab:noise_types}
    \begin{tabular}{l|p{0.55\textwidth}|p{0.32\textwidth}}
        \toprule
        \textbf{Noise Type} & \textbf{Description} & \textbf{Example} \\
        \midrule
        Slang & Replace formal words with slang or informal abbreviations & "I don't know" → "idk" \\
        \midrule
        Case & Change one character from lowercase to uppercase or vice versa & "my card" → "My card" \\
        \midrule
        Pronoun & Replace nouns with pronouns & "my card" → "this" \\
        \midrule
        Replace Word & Replace words with semantically similar alternatives & "purchase" → "buy" \\
        \midrule
        Delete Word & Remove a word from the sentence & "I want my card" → "I want card" \\
        \midrule
        Swap Words & Exchange the positions of two adjacent words & "my card" → "card my" \\
        \midrule
        Homophone & Replace words with phonetically similar alternatives & "there" → "their" \\
        \midrule
        Cyrillic & Replace Latin characters with visually similar Cyrillic characters & "like" → "li\underline{{\foreignlanguage{russian}{к}}}e" \\
        \midrule
        Keyboard-char & Replace a character with a neighboring key on the keyboard (common typo) & "time" → "rime" \\
        \bottomrule
    \end{tabular}
\end{table*}
\FloatBarrier
\section{Additional Results}
\label{Additional_Results}

\FloatBarrier

\subsection{Token-Breaking Examples}
\label{break_token}

A \emph{token-breaking} edit is any character-level change that forces the
tokenizer to split what was a single token into several fragments.  
The table below shows three such edits.

\begin{center}
\begin{tabular}{@{}lcc@{}}
\toprule
\textbf{String} & \textbf{Tokenizer output} & \textbf{\#} \\
\midrule
reactivate & \texttt{\_reactivate}         & 1 \\
reactviate & \texttt{\_react\ via\ te}     & 3 \\[2pt]
card       & \texttt{\_card}              & 1 \\
crd        & \texttt{\_c\ rd}             & 2 \\[2pt]
would      & \texttt{\_would}             & 1 \\
woudd      & \texttt{\_wo\ ud\ d}         & 3 \\
\bottomrule
\end{tabular}
\end{center}

\begin{itemize}\setlength\itemsep{2pt}
  \item A single, well-trained token is replaced by several rare ones; their
        embeddings lie far from the original word in representation space.
  \item The longer sequence alters positional patterns, further distancing the
        whole sentence from its clean counterpart.
  \item Because these new embeddings sit \emph{outside} the uncertainty region
        sampled during adversarial training, the defence becomes much less
        effective against such edits.
\end{itemize}

These observations explain the large gap between character-noise points
(triangles) and word-level/clean points (circles, squares) in
\cref{fig:trec_noise}.

\FloatBarrier
\clearpage


\begin{table*}[!b]
\centering
\small
\caption{\textbf{Adversarial Training Improves PEFT Across Resource Settings.} Performance comparison of standard PEFT methods vs their adversarial training counterparts (SDBN) on text classification tasks using BERT-base. Results show accuracy ({\%}) when training on varying percentages (100\%, 50\%, 20\%, 10\%, 5\%) of each dataset. Bold numbers indicate the better performing variant for each method pair. Adversarial training improves classification performance across all PEFT methods, datasets, and resource settings, with particularly significant gains in low-resource scenarios.}
\label{tab:all_scales}
\setlength{\tabcolsep}{4pt}
\begin{tabular}{ll|cccc}
\toprule
\textbf{Data \%} & \textbf{Method} & \textbf{Banking77} & \textbf{TREC} & \textbf{20NewsGroups} & \textbf{IMDB} \\
\midrule
100\% & Adapter & 92.82$\pm$0.40 & 91.16$\pm$0.39 & 68.67$\pm$0.52 & 87.71$\pm$0.33 \\
& \textbf{SDBN-Adapter} & \textbf{93.29$\pm$0.10} & \textbf{91.88$\pm$0.56} & \textbf{69.62$\pm$0.49} & \textbf{88.96$\pm$0.20} \\
\cmidrule{2-6}
& BitFit & \textbf{90.67$\pm$0.45} & \textbf{90.20$\pm$0.55} & 66.86$\pm$0.46 & 87.03$\pm$0.95 \\
& \textbf{SDBN-BitFit} & 90.66$\pm$0.34 & 89.96$\pm$0.81 & \textbf{67.57$\pm$0.29} & \textbf{87.87$\pm$1.07} \\
\cmidrule{2-6}
& LoRA & 91.56$\pm$0.18 & 91.44$\pm$1.04 & 68.84$\pm$0.27 & 88.30$\pm$0.13 \\
& \textbf{SDBN-LoRA} & \textbf{91.77$\pm$0.51} & \textbf{92.12$\pm$0.16} & \textbf{69.60$\pm$0.25} & \textbf{88.89$\pm$0.26} \\
\midrule
50\% & Adapter & 90.58$\pm$0.32 & 90.12$\pm$0.68 & 66.93$\pm$0.36 & 86.68$\pm$0.43 \\
& \textbf{SDBN-Adapter} & \textbf{91.41$\pm$0.17} & \textbf{90.72$\pm$0.79} & \textbf{67.80$\pm$0.22} & \textbf{87.97$\pm$0.23} \\
\cmidrule{2-6}
& BitFit & 87.67$\pm$0.40 & 88.08$\pm$0.48 & 65.46$\pm$0.40 & 86.84$\pm$0.25 \\
& \textbf{SDBN-BitFit} & \textbf{88.39$\pm$0.30} & 88.08$\pm$0.52 & \textbf{65.89$\pm$0.41} & \textbf{87.46$\pm$0.11} \\
\cmidrule{2-6}
& LoRA & 89.05$\pm$0.23 & \textbf{90.20$\pm$1.15} & 66.42$\pm$0.31 & 87.52$\pm$0.14 \\
& \textbf{SDBN-LoRA} & \textbf{89.92$\pm$0.40} & 90.16$\pm$0.67 & \textbf{67.85$\pm$0.23} & \textbf{87.98$\pm$0.16} \\
\midrule
20\% & Adapter & 86.18$\pm$0.69 & 83.68$\pm$1.26 & 64.64$\pm$0.45 & 85.52$\pm$0.42 \\
& \textbf{SDBN-Adapter} & \textbf{87.20$\pm$0.76} & \textbf{84.40$\pm$0.99} & \textbf{65.33$\pm$0.45} & \textbf{86.70$\pm$0.15} \\
\cmidrule{2-6}
& BitFit & 79.59$\pm$0.87 & 78.40$\pm$1.01 & 61.61$\pm$0.73 & 85.13$\pm$0.75 \\
& \textbf{SDBN-BitFit} & \textbf{81.61$\pm$0.86} & \textbf{79.60$\pm$0.97} & \textbf{62.70$\pm$0.70} & \textbf{86.44$\pm$0.16} \\
\cmidrule{2-6}
& LoRA & 82.47$\pm$0.75 & 81.88$\pm$1.83 & 63.17$\pm$0.50 & 86.21$\pm$0.26 \\
& \textbf{SDBN-LoRA} & \textbf{84.81$\pm$0.75} & \textbf{83.24$\pm$1.72} & \textbf{64.28$\pm$0.50} & \textbf{86.56$\pm$0.23} \\
\midrule
10\% & Adapter & 78.26$\pm$0.96 & 74.40$\pm$1.93 & 63.16$\pm$0.37 & 84.78$\pm$0.30 \\
& \textbf{SDBN-Adapter} & \textbf{80.48$\pm$0.97} & \textbf{76.76$\pm$1.53} & \textbf{63.47$\pm$0.45} & \textbf{85.39$\pm$0.17} \\
\cmidrule{2-6}
& BitFit & 67.45$\pm$1.31 & 65.68$\pm$1.83 & 56.78$\pm$1.38 & 84.01$\pm$0.35 \\
& \textbf{SDBN-BitFit} & \textbf{70.08$\pm$0.71} & \textbf{69.24$\pm$0.80} & \textbf{58.92$\pm$0.92} & \textbf{84.56$\pm$0.95} \\
\cmidrule{2-6}
& LoRA & 66.42$\pm$1.97 & 72.20$\pm$1.57 & 60.13$\pm$0.75 & 85.46$\pm$0.31 \\
& \textbf{SDBN-LoRA} & \textbf{72.86$\pm$1.24} & \textbf{74.72$\pm$1.43} & \textbf{61.56$\pm$0.53} & \textbf{86.00$\pm$0.15} \\
\midrule
5\% & Adapter & 58.69$\pm$1.72 & 59.32$\pm$4.29 & 59.74$\pm$1.04 & 84.08$\pm$0.37 \\
& \textbf{SDBN-Adapter} & \textbf{63.70$\pm$1.50} & \textbf{63.92$\pm$3.85} & \textbf{61.07$\pm$1.00} & \textbf{84.89$\pm$0.43} \\
\cmidrule{2-6}
& BitFit & 44.30$\pm$1.81 & 51.96$\pm$1.61 & 52.29$\pm$0.96 & 82.51$\pm$0.66 \\
& \textbf{SDBN-BitFit} & \textbf{48.21$\pm$2.10} & \textbf{56.32$\pm$3.66} & \textbf{53.88$\pm$0.70} & \textbf{83.64$\pm$0.90} \\
\cmidrule{2-6}
& LoRA & 36.29$\pm$3.89 & 56.04$\pm$5.03 & 55.86$\pm$2.06 & 84.63$\pm$0.25 \\
& \textbf{SDBN-LoRA} & \textbf{43.30$\pm$2.57} & \textbf{61.76$\pm$1.64} & \textbf{58.27$\pm$0.93} & \textbf{85.22$\pm$0.22} \\
\bottomrule
\end{tabular}

\end{table*}

\label{results_clean}
\newpage

\FloatBarrier
\subsection{Performance over noisy test set}
\label{app::noise_const}
\begin{figure}[h]
  \centering
  \tiny
  \includegraphics[width=0.5\textwidth]{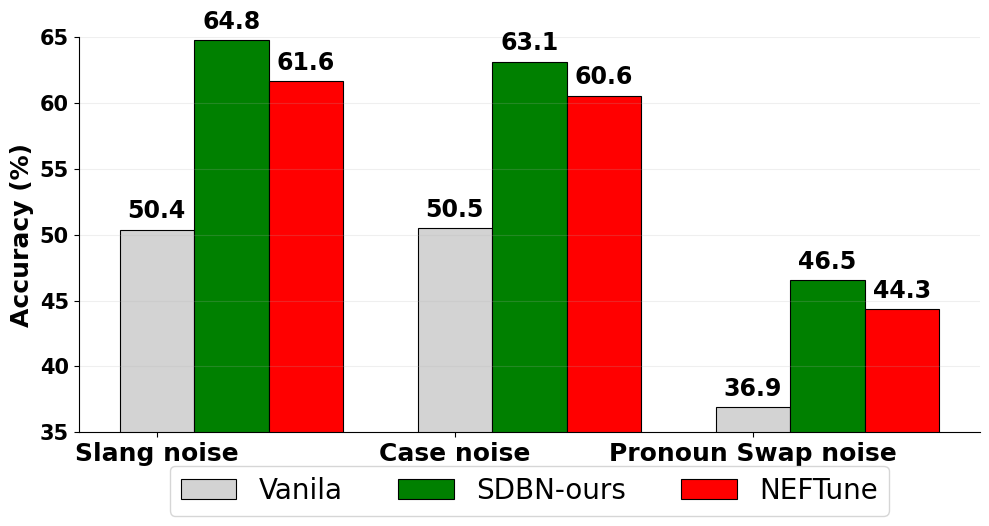}
  \caption{\textbf{Adversarial training for constant-intensity noise.}
  Performance comparison of BitFit PEFT implementations under constant-intensity
  noise conditions with DeBERTa-v3 on 1,000 samples from the \textit{Banking77}
  dataset. These results demonstrate the superior robustness of gradient-based
  adversarial training in low-resource settings.}
\label{fig::noise_const_bitfit}
\end{figure}

\textbf{Constant-Intensity Noise Results.} Figure~\ref{fig::noise_const_bitfit} presents results for constant-intensity perturbations, where exactly one operation (word deletion, swap, replacement, etc.) is applied per sentence. SDBN consistently outperforms both vanilla PEFT and NEFTune across all noise types, with improvements ranging from 3-8\% absolute accuracy. These single-operation perturbations represent the minimal corruptions commonly found in real-world text, demonstrating SDBN's effectiveness even for subtle linguistic variations.

\label{appendix_noise_res}

\begin{figure}[h]
\centering
  \includegraphics[width=0.5\textwidth]{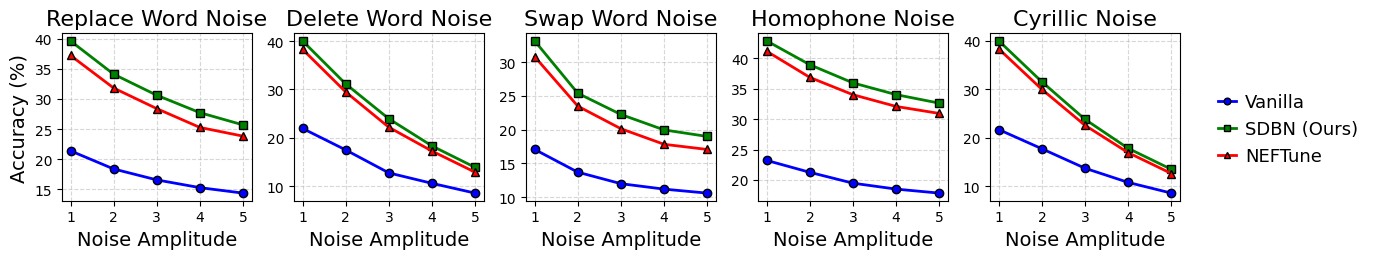}
\caption{\textbf{Adversarial training for variable-intensity noise, Adapter PEFT.} Performance comparison of Adapter PEFT implementations under variable-intensity noise conditions in different amplitudes with DeBERTa-v3 on 1,000 samples from Banking77 dataset. These results demonstrate the superior robustness of Adversarial training gradient-based to noise in low-resource settings.}
\end{figure}

\FloatBarrier

\begin{figure}[ht]
\centering
  \includegraphics[width=0.5\textwidth]{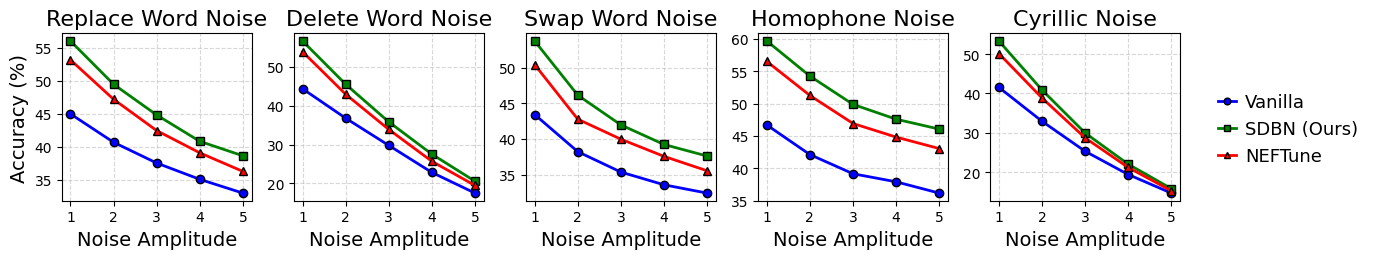}
\caption{\textbf{Adversarial training for variable-intensity noise, BitFit PEFT.} Performance comparison of BitFit PEFT implementations under variable-intensity noise conditions in different amplitudes with DeBERTa-v3 on 1,000 samples from Banking77 dataset. These results demonstrate the superior robustness of Adversarial training gradient-based to noise in low-resource settings.}
\end{figure}

\begin{figure}[h]
\centering
  \includegraphics[width=0.5\textwidth]{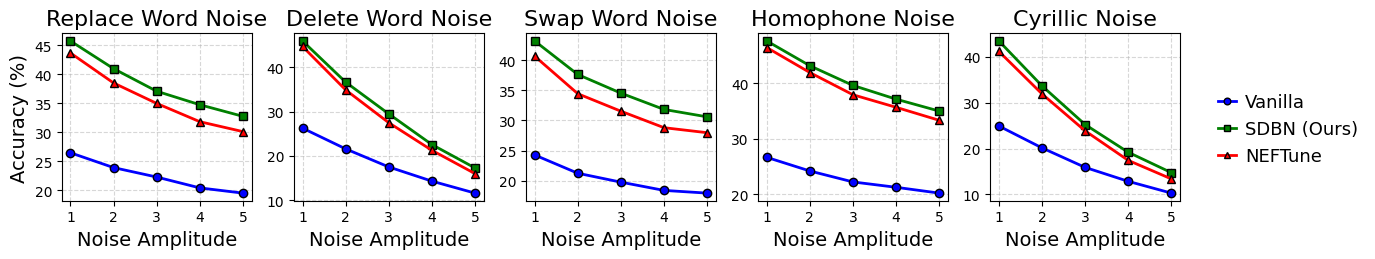}
\caption{\textbf{Adversarial training for variable-intensity noise, QLoRA PEFT.} Performance comparison of QLoRA PEFT implementations under variable-intensity noise conditions in different amplitudes with DeBERTa-v3 on 1,000 samples from Banking77 dataset. These results demonstrate the superior robustness of Adversarial training gradient-based to noise in low-resource settings.}
\end{figure}
\FloatBarrier

\begin{figure}[H]
\includegraphics[width=0.5\textwidth]{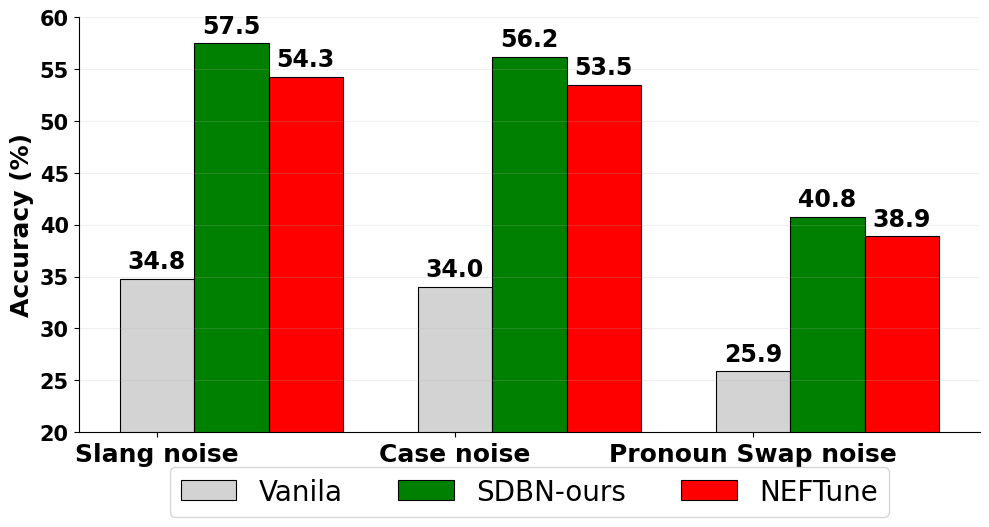}
\caption{\textbf{Adversarial training for constant-intensity noise, LoRA PEFT.} Performance comparison of LoRA PEFT implementations under constant-intensity noise conditions with DeBERTa-v3 on 1,000 samples from Banking77 dataset. These results demonstrate the superior robustness of Adversarial training gradient-based to noise in low-resource settings.}
\end{figure}

\begin{figure}[H]
\includegraphics[width=0.5\textwidth]{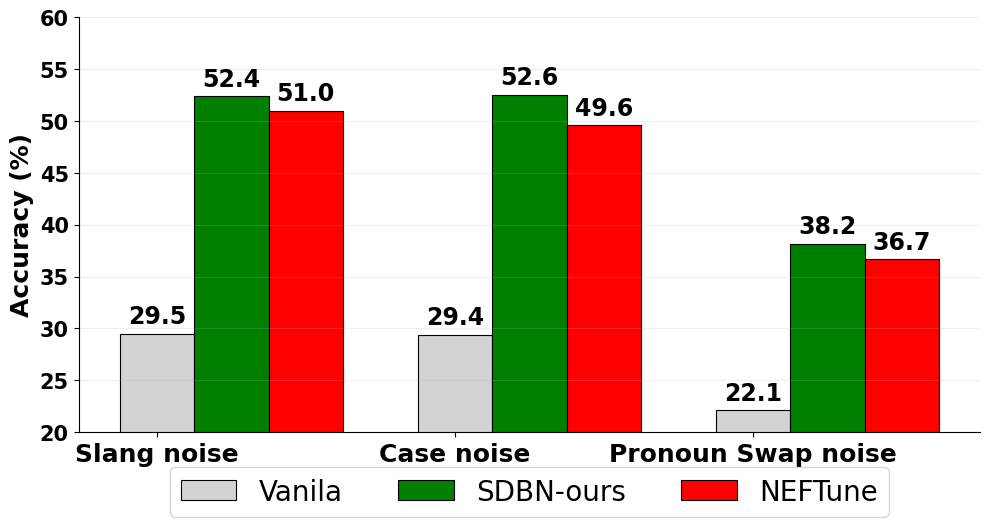}
\caption{\textbf{Adversarial training for constant-intensity noise, QLoRA PEFT.} Performance comparison of QLoRA PEFT implementations under constant-intensity noise conditions with DeBERTa-v3 on 1,000 samples from Banking77 dataset. These results demonstrate the superior robustness of Adversarial training gradient-based to noise in low-resource settings.}
\end{figure}

\FloatBarrier
\begin{figure}[H]
\includegraphics[width=0.5\textwidth]{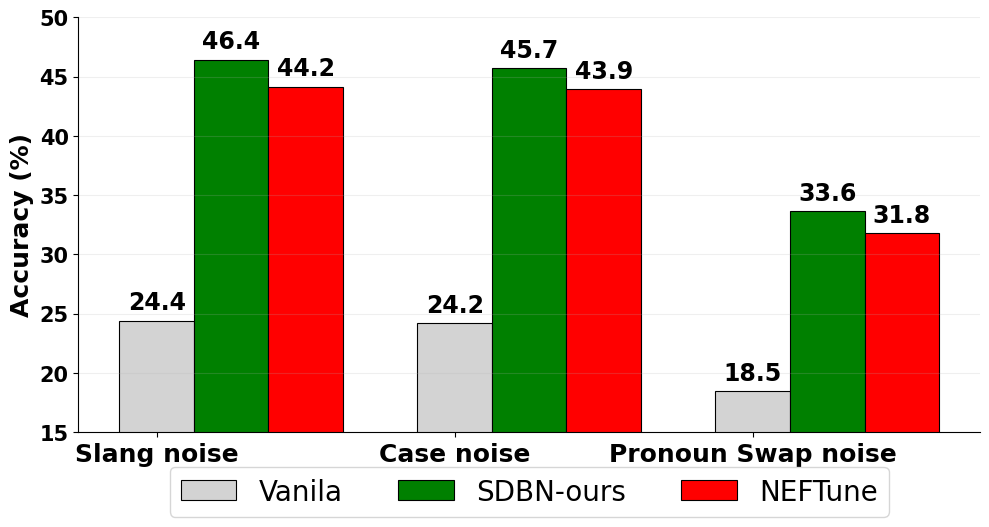}
\caption{\textbf{Adversarial training for constant-intensity noise, Adapter PEFT.} Performance comparison of Adapter PEFT implementations under constant-intensity noise conditions with DeBERTa-v3 on 1,000 samples from Banking77 dataset. These results demonstrate the superior robustness of Adversarial training gradient-based to noise in low-resource settings.}
\end{figure}

\FloatBarrier

\begin{figure}[ht]
  \centering
  \includegraphics[width=\linewidth]{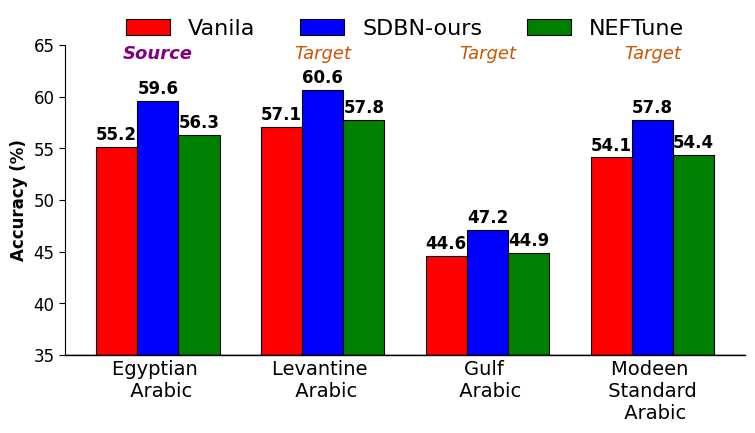}
  \caption{\textbf{Domain Shifts.} Comparison of LoRA PEFT methods on Arabic
  dialect sentiment classification using multilingual BERT-base trained on 270 samples
  of Egyptian Arabic without exposure to the other dialects during training.
  SDBN consistently outperforms alternatives across all dialects, with notable
  advantages on both the source and target domains.}
  \label{fig:domain_shifts}
\end{figure}
\label{app:domain-shift} 
\FloatBarrier
\begin{figure}[H]
    \centering
    \includegraphics[width=0.5\textwidth]{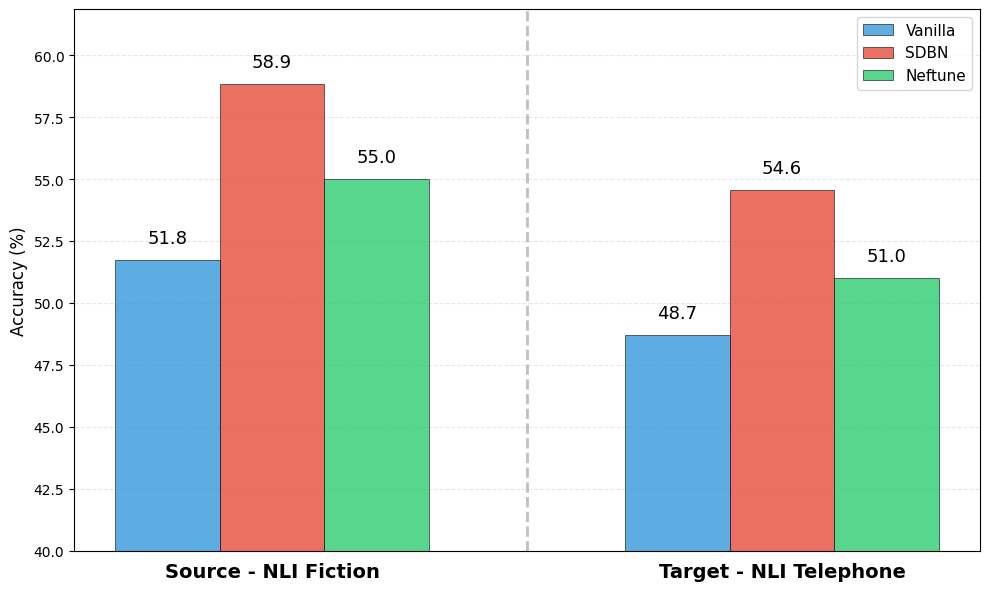}
    \caption{\textbf{NLI Classification Performance: Source vs Target Domains.} Performance comparison of LoRA PEFT methods on NLI  classification tasks using DeBERTa-v3-large trained on only 77 samples from the source domain (NLI Fiction) without exposure to the target domain during training. Results show mean classification accuracy across 20 random seeds. SDBN consistently outperforms both vanilla LoRA and NEFTune across both domains, demonstrating superior robustness and generalization in this text classification scenario. While NEFTune shows improvement over vanilla LoRA through random noise injection, SDBN's gradient-based adversarial perturbations provide more robustness, particularly valuable in this extreme low-resource classification setting.}
    \label{fig:domain_shifts_nli}
\end{figure}
\FloatBarrier
\subsection{Domain Shifts.}
\textbf{Domain Shifts.}\label{sec:domain_shifts} In a \emph{low-resource} setup- 270 training sentences from a single Arabic dialect and \emph{no exposure to the target dialect}-we evaluate cross-domain robustness on the ArSarcasm-v2 benchmark. Despite this tiny-data constraint \cref{fig:domain_shifts}, SDBN generalizes better to the unseen dialect than both vanilla and NEFTune.

This improvement accords with the rationale in \cref{sec::rationale}: gradient-based perturbations explore uncertainty regions around each source example that extend beyond the source dialect.  
The t-SNE visualization in \cref{fig:tsne_domain} shows these regions overlapping semantically related sentences from the unseen dialect, effectively “covering” the target domain without additional supervision.

Optimizing on these worst-case neighbors equips SDBN to handle domain shifts that would otherwise demand costly labeled data--an advantage that becomes particularly decisive when the available corpus is as small as a few hundred examples.

See \cref{app:domain-shift} for further experiments and results on domain shifts.

\subsection{Generative Tasks}
\label{generative_tasks_appendix}
\begin{table}[H]
  \centering
  \caption{\textbf{SQuAD generative QA results (EM / F1).}
    Models trained on 500 clean SQuAD examples with DeBERTa-v3 and tested on clean data and noisy variants.
    Two noise types at test time: \textsc{Delete-Char} and \textsc{Delete-Word}.
    Results are averaged over 5 runs. SDBN outperforms both vanilla PEFT and NEFTune.}
  \label{tab:squad_results}
  \small
  \setlength{\tabcolsep}{3pt}  
  \begin{tabular}{l@{\hspace{6pt}}c@{\hspace{6pt}}c@{\hspace{6pt}}c}
    \toprule
    & \textbf{Clean} & \textbf{Del-Char} & \textbf{Del-Word} \\
    \midrule
    Vanilla & 55.0/69.8 & 51.6/66.2 & 49.3/63.9 \\
    SDBN & \textbf{57.8/72.2} & \textbf{54.1/68.6} & \textbf{51.7/66.3} \\
    NEFTune & 55.0/69.6 & 51.5/66.1 & 48.9/63.5 \\
    \bottomrule
  \end{tabular}
\end{table}

\begin{table}[h]
\centering
\small
\caption{Additional generative results. TweetQA F1 using LLaMA-3.2-1B with LoRA, trained on 200 samples. Reported values are mean{\scriptsize$\pm$std} over random seeds. Best results in each column are shown in \textbf{bold}; second-best results are \underline{underlined}.}
\label{tab:tweetqa_llama1b}
\begin{tabular}{lccc}
\toprule
Method & Clean & Keyboard-Char & Cyrillic \\
\midrule
Vanilla  & 65.5{\scriptsize$\pm$1.2}             & 43.2{\scriptsize$\pm$1.7}             & 26.8{\scriptsize$\pm$1.3} \\
EDA      & 65.1{\scriptsize$\pm$3.0}             & \underline{45.3{\scriptsize$\pm$2.7}} & \underline{33.4{\scriptsize$\pm$3.0}} \\
NEFTune  & \underline{66.9{\scriptsize$\pm$1.3}} & 44.7{\scriptsize$\pm$0.6}             & 33.0{\scriptsize$\pm$2.8} \\
SMART    & 66.6{\scriptsize$\pm$0.6}             & 43.9{\scriptsize$\pm$2.0}             & 21.2{\scriptsize$\pm$3.2} \\
SDBN-p   & \textbf{67.2{\scriptsize$\pm$2.8}}    & \textbf{50.6{\scriptsize$\pm$2.0}}    & \textbf{41.2{\scriptsize$\pm$0.9}} \\
\bottomrule
\end{tabular}
\end{table}

\begin{table}[h]
\centering
\small
\caption{Additional generative results. SQuAD exact match (EM) using Qwen-2.5-7B with LoRA, trained on 200 samples. Reported values are mean{\scriptsize$\pm$std} over random seeds. Best results in each column are shown in \textbf{bold}; second-best results are \underline{underlined}.}
\label{tab:squad_qwen7b}
\begin{tabular}{lccc}
\toprule
Method & Clean & Keyboard-Char & Double-Char \\
\midrule
Vanilla  & 58.00{\scriptsize$\pm$3.00}             & 47.44{\scriptsize$\pm$3.13}             & 48.28{\scriptsize$\pm$3.63} \\
EDA      & \underline{59.48{\scriptsize$\pm$1.88}} & \underline{50.08{\scriptsize$\pm$2.29}} & \underline{50.32{\scriptsize$\pm$2.28}} \\
NEFTune  & 58.12{\scriptsize$\pm$2.76}             & 47.92{\scriptsize$\pm$3.35}             & 48.60{\scriptsize$\pm$3.49} \\
FreeLB   & 53.92{\scriptsize$\pm$1.81}             & 49.32{\scriptsize$\pm$2.74}             & 49.68{\scriptsize$\pm$2.64} \\
SDBN-p   & \textbf{72.84{\scriptsize$\pm$4.18}}    & \textbf{69.04{\scriptsize$\pm$3.82}}    & \textbf{69.52{\scriptsize$\pm$3.11}} \\
\bottomrule
\end{tabular}
\end{table}

For \textbf{SQuAD}, we adopt a compact adversarial-training schedule consisting of 1 warm-up epoch followed by 10 adversarial/method-specific epochs. For each training example, we construct 5 variants, and the same fixed set of variants is reused throughout training. 

For \textbf{TweetQA}, we use a longer and more dynamic training schedule with 5 warm-up epochs followed by 16 adversarial/method-specific epochs. For each training example, we generate 10 variants; unlike SQuAD, these variants are regenerated at every adversarial epoch, so each epoch exposes the model to a new set of perturbations. This dynamic setup provides more diverse adversarial supervision and broader coverage of linguistic variation across training.

\subsection{Method Analysis}
\label{sec:method_analysis}

We analyze the primary implementation choices that underlie the effectiveness of our approach.  
First, we compared perturbations constrained by the $\ell_{1}$, $\ell_{2}$, and $\ell_{\infty}$ norms across a range of magnitudes.  
The $\ell_{\infty}$ norm with $\epsilon = 10^{-4}$ produced the best performance over other norms.  
Full details of the norm comparison and $\epsilon$ tuning appear in \cref{eps_exp_apendix}.

Next, we explored \emph{where} to inject the perturbations along the Transformer stack.  
When the noise was added to hidden activations in intermediate encoder layers, accuracy deteriorated sharply.  
Restricting the perturbation to the \textit{embedding layer}, however, preserved the model’s semantic processing and yielded the strongest robustness gains.  
Applying perturbations solely at the \textit{embedding layer} keeps the noise
interpretable as realistic textual edits--such as swapping two words, inserting
an extra token, or deleting a character--thereby exposing the model to linguistic variation.  Injecting noise deeper in the encoder, by contrast,
corrupts high-level semantic representations that no valid sentence could
produce, which explains the strong performance drop we observe.  Detailed
results for every injection point are reported in \cref{injected_layer}.


\begin{table}[ht]
  \centering
  \footnotesize
  \caption{\textbf{Runtime per batch (ms).} Mean wall-clock time (milliseconds) on an NVIDIA H200
  for a full training step (forward--backward) on a batch of 32 \textit{Banking77} sentences across PEFT methods.
  Lower is faster.}
  \label{tab:runtime}
  \begin{tabular}{lrrrr}
    \toprule
    \textbf{Approach} & \textbf{LoRA} & \textbf{QLoRA} & \textbf{BitFit} & \textbf{Adapter} \\
    \midrule
    Vanilla  & 116.21 & 148.96 &  92.68 &  95.79 \\
    SDBN     & 163.05 & 189.70 & 123.40 & 131.22 \\
    SDBN-h   & 170.71 & 206.64 & 140.48 & 150.25 \\
    SDBN-p   & 211.91 & 247.98 & 171.87 & 178.85 \\
    \bottomrule
  \end{tabular}
\end{table}

\paragraph{Runtime.}
Table~\ref{tab:runtime} indicates that SDBN variants introduce a \emph{bounded} and \emph{predictable} compute overhead relative to vanilla PEFT. 
SDBN increases per-batch latency by about \(1.40\times\), \(1.27\times\), \(1.33\times\), and \(1.37\times\) for LoRA, QLoRA, BitFit, and Adapter, respectively (e.g., LoRA: \(116.21\!\to\!163.05\) ms). 
The hybrid SDBN-h adds only a modest margin over SDBN—approximately 5–15\% across methods (LoRA: \(163.05\!\to\!170.71\) ms; QLoRA: \(189.70\!\to\!206.64\) ms; BitFit: \(123.40\!\to\!140.48\) ms; Adapter: \(131.22\!\to\!150.25\) ms). 
SDBN-p incurs a larger but still bounded overhead—approximately 30–39\% over SDBN (LoRA: \(163.05\!\to\!211.91\) ms; QLoRA: \(189.70\!\to\!247.98\) ms; BitFit: \(123.40\!\to\!171.87\) ms; Adapter: \(131.22\!\to\!178.85\) ms). 
Overall, the added cost remains practical for deployment: the methods retain parameter efficiency and deliver the robustness gains reported earlier while keeping per-batch latency within practical bounds.

\begin{table}[h]
\centering
\small
\begin{tabular}{lc}
\toprule
\textbf{Method} & \textbf{Peak Memory (GiB)} \\
\midrule
Vanilla & 68.60 \\
SDBN & 68.70 \\
SDBN-h & 68.70 \\
SDBN-p & 68.60 \\
\bottomrule
\end{tabular}
\caption{Peak GPU memory usage during training on H200.}
\label{tab:memory}
\end{table}

\textbf{Memory Efficiency.}
Table~\ref{tab:memory} reports peak GPU memory during training. All SDBN variants introduce negligible memory overhead ($\leq$0.15\%) compared to vanilla fine-tuning, confirming that our framework enhances robustness without additional memory cost. Notably, SDBN-p requires no extra memory since adversarial variants are pre-computed offline.

\begin{table*}[!h]
  \centering
  \small
  \setlength{\tabcolsep}{6pt}
  \renewcommand{\arraystretch}{1.12}
  \caption{\textbf{A qualitative example.}
  A DeBERTa-v3 encoder was fine-tuned with LoRA on 1{,}000 \emph{clean} \textsc{Banking77} training examples.
  We show one clean test example (“Original”) and several noisy variants (word- and character-level).
  Cells report correctness (\checkmark\ = correct, $\times$ = incorrect).
  Adversarial training improves robustness overall (SDBN), and its hybrid variant (SDBN-h) is especially effective on character-level noise that disrupts tokenization.}
  \label{tab:qual}
  \begin{tabularx}{\textwidth}{@{}l Y C C C C@{}}
    \toprule
    \textbf{Noise type} & \textbf{Example} & \textbf{SDBN} & \textbf{SDBN-h} & \textbf{Vanilla} & \textbf{NEFTune} \\
    \midrule
    Original
      & \emph{How do I re-add a card to the app?}
      & \checkmark & \checkmark & \checkmark & \checkmark \\
    Case char
      & \emph{\textbf{h}Ow \textbf{D}o \textbf{i} \textbf{Re}-\textbf{aDd} A \textbf{caRd} \textbf{tO} \textbf{tHe} app?}
      & $\times$ & \checkmark & $\times$ & $\times$ \\
    Replace word
      & \emph{How do \textbf{me} \textbf{reinsert} a \textbf{ticket} to the \textbf{tool}?}
      & \checkmark & \checkmark & \checkmark & $\times$ \\
    Insert char
      & \emph{H\textbf{1}o\textbf{2}w d\textbf{3}o I r\textbf{4}e-add a card to the app?}
      & $\times$ & \checkmark & $\times$ & $\times$ \\
    Delete word
      & \emph{How do I re-add\textbf{ \ldots}?} 
      & \checkmark & \checkmark & $\times$ & \checkmark \\
    Delete char
      & \emph{How do I re-\textbf{dd} a card \textbf{t} the app?}
      & \checkmark & \checkmark & $\times$ & $\times$ \\
    \bottomrule
  \end{tabularx}
\end{table*}

\subsection{A Qualitative Example.}
\label{qualitive_example}
\Cref{tab:qual} demonstrates that gradient-based adversarial training (SDBN) reliably improves robustness on word-level edits over Vanilla and NEFTune while preserving clean-data accuracy.
Consistent with our rationale in \cref{sec::rationale}, optimizing against worst-case neighbors teaches the model to correctly handle semantically preserving rewrites that lie within the local uncertainty set.
Importantly, the hybrid variant SDBN-h which augments SDBN with discrete character perturbations—substantially mitigates tokenization-breaking edits (character changes), closing the gap on character-level noise that defeats both baselines.
Overall, these examples indicate that SDBN strengthens robustness around clean inputs, and SDBN-h extends this robustness to fine-grained character corruptions, yielding the most consistent behavior across the noisy variants in \Cref{tab:qual}.

\subsection{Setup Details}
\label{setup_details}

For all experiments, we utilized the Hugging Face \texttt{bert-base} model as our pre-trained backbone. The implementation of Parameter-Efficient Fine-Tuning (PEFT) methods was conducted using the PEFT library with PyTorch.
We employed the AdamW optimizer with a learning rate of $1\times10^{-4}$, applied only to trainable parameters.
The LoRA rank was set to 12 for BERT models and 4 for DeBERTa models across both the baseline and our adversarial training method. The adapter bottleneck dimension was set to 16 for both methods.
Each experimental setting (dataset-method-dataset size) was run five times with five different random seeds to ensure robustness. During fine-tuning, we used:
Batch size: 32
Dropout: 0.1
Evaluation was performed using the TextAttack library to assess model noise robustness.
All experiments were conducted using an NVIDIA GPU. In \cref{results_clean} results show mean accuracy across 5 random seeds. In \cref{appendix_noise_res} results show mean accuracy across 10 random seeds without outliers (excluding the best and worst results for each method to reduce bias when working with small datasets).

\paragraph{EDA baseline.}
\label{app::eda}
We use EDA as a lightweight data-noising baseline.
To keep training compute comparable, we use conservative augmentation rates
($\alpha_{\text{sr}}{=}0.05$, $\alpha_{\text{ri}}{=}0.05$, $\alpha_{\text{rs}}{=}0.05$, $p_{\text{rd}}{=}0.05$)
and generate a single augmented example per input ($\texttt{num\_aug}{=}1$).
To match the training compute of other methods (i.e., the same number of optimizer steps),
we do not expand the dataset; instead, in each mini-batch we replace the clean input with its EDA-augmented variant
with probability $0.5$, following the original EDA procedure. Note that EDA was originally proposed as a general data augmentation method and was not presented in combination with PEFT in its original work. The augmented dataset is generated once before training and reused across all epochs, following the original EDA offline augmentation procedure

\end{document}